%% file: emnlp2026_draft.tex
\documentclass[11pt]{article}
\usepackage[preprint]{acl}

\usepackage{times}
\usepackage{latexsym}
\usepackage[T1]{fontenc}
\usepackage[utf8]{inputenc}
\usepackage{microtype}
\usepackage{inconsolata}
\usepackage{graphicx}
\usepackage{booktabs}
\usepackage{amsmath}
\usepackage{amssymb}
\usepackage{multirow}
\usepackage{tabularx}
\usepackage{fvextra}
\usepackage{array}
\usepackage{bm}
\DefineVerbatimEnvironment{PromptBox}{Verbatim}{
  breaklines=true,
  breakanywhere=true,
  fontsize=\small
}
\usepackage{soul, xcolor}

\title{ER-EDF: A Psychology-Grounded Emotion Regulation Framework for Speech Empathetic Dialogue Generation in Large Audio-Language Models}

\author{Hongyu Jin\textsuperscript{1} \quad Wenda Zhang\textsuperscript{1} \quad Runqiu Fei\textsuperscript{1} \\
\textbf{Gongping Huang\textsuperscript{2} \quad Mike Conway\textsuperscript{1} \quad Ting Dang\textsuperscript{1}\thanks{Corresponding author: \texttt{ting.dang@unimelb.edu.au}.}} \\
\textsuperscript{1}The University of Melbourne, Australia \\
\textsuperscript{2}Wuhan University, China \\
\small\texttt{hongyuj1@student.unimelb.edu.au}
}

\begin{document}
\maketitle

\begin{abstract}
In human conversation, empathy involves not only recognizing a partner's emotion but also regulating it, listeners identify the emotional state, select a suitable regulation strategy, and implement it in their reply. Current large audio–language models (LALMs) largely map perceived emotion directly onto a response, skipping this regulation process. As a result, their responses tend to mirror the user’s affect or offer generic sympathy.  %
We propose ER-EDF, a psychology-grounded framework for empathetic dialogue
that follows the three stages of Gross's Extended Process Model of emotion
regulation, identifying the user's emotion from
speech, selecting a regulation strategy matched to that emotion, and
implementing the strategy through a LALM. The framework is model-agnostic and integrates seamlessly into existing LALMs. We further construct a spoken empathetic dialogue dataset and introduce empathy-aware evaluation metrics beyond lexical matching. Experiments across five LALMs and two datasets show that ER-EDF consistently improves empathetic response quality in both automatic and human evaluations. These results suggest that modeling how humans regulate others' emotions, rather than only how they perceive them, is key to empathetic spoken dialogue, paving the way for psychologically grounded empathetic AI.
\end{abstract}

\input{latex/sections/intro}
\input{latex/sections/related_work}
\input{latex/sections/method}

\input{latex/sections/Experiments}
\input{latex/sections/results}

\input{latex/sections/conclusion}
\section{Limitations}

This work opens several directions for future improvement. First, although we validate the constructed empathetic references with both automatic metrics and human judgments, broader human evaluation would further strengthen the reliability of the analysis. The current evaluation provides useful evidence for reference quality and model preference, but future studies can benefit from larger sample coverage, more annotators, and more fine-grained criteria, such as emotional appropriateness, conversational coherence, perceived helpfulness, and regulation effectiveness. Such expanded evaluation would provide a more comprehensive understanding of how users perceive empathetic responses in different dialogue contexts.

Second, the current emotion regulation policy in ER-EDF is predefined according to emotion regulation theory and valence-arousal rules. This design makes the framework interpretable and easy to audit, which is important for affect-sensitive dialogue systems. A promising future direction is to extend this rule-based policy into a learned dynamic regulation module. Such a module could infer regulation goals and response strategies from richer conversational signals, including speaker identity, dialogue history, pragmatic intent, and the user's reaction to previous responses, while still preserving the interpretability of the regulation decision.

Third, ER-EDF currently uses a discrete valence-arousal mapping to unify heterogeneous emotion labels across datasets. This provides a simple and dataset-agnostic affective representation, enabling consistent trajectory tracking and regulation planning. Future work could further enrich this representation by learning continuous valence-arousal estimates through regression. A continuous two-dimensional affect space would allow the framework to capture finer-grained differences in emotional intensity, ambiguity, and mixed affect, supporting more precise trajectory modeling and more adaptive regulation-aware response generation.

\bibliography{custom}
\newpage
\appendix
\section{appendix}
\input{latex/appendix/prompts}
\input{latex/appendix/method_app}
\input{latex/appendix/SER}
\input{latex/appendix/experiment_settings}
\input{latex/appendix/results_app}

\input{latex/appendix/human}
\end{document}

%% file: latex/sections/intro.tex
\section{Introduction}

The rapid advancement of spoken conversational AI systems, including ChatGPT~\citep{openai2024gpt4o} and Google Gemini~\citep{google2024gemini}, has established a new paradigm of natural spoken human–computer interaction. Beyond task completion, these systems increasingly support affect-sensitive applications such as mental health support, eldercare companionship, and emotional well-being coaching~\citep{laranjo2018conversational, sharma2020computational}. Such settings require models to perceive and appropriately respond to users’ emotional states. Empathetic response generation is therefore a core requirement, and its failure can undermine trust and pose risks in vulnerable contexts.

Responding empathetically requires more than recognizing a partner's emotion~\citep{preston2002empathy}; it requires helping them regulate it. 
Interpersonal emotion regulation research shows that listeners actively shape a partner's affect through validation, calming, or shared positive engagement~\citep{zaki2013interpersonal,niven2009classification}. Gross's Extended Process Model~\citep{Gross2015-bk} formalizes this as three stages: identifying whether and what regulation is needed, selecting a strategy, and implementing it. Current LALMs, however, have no such regulation mechanism,  collapsing complex human-being empathy into directly perceived emotion
As a result, they tend to mirror the user's affect or express generic empathy even when the user's state does not call for it. 
Recent LALM-based systems have begun to address empathetic spoken dialogue in two ways. Some strengthen speech emotion recognition~\citep{xue2024echat,lin2024paralingpt,wang2024blspemo} to condition the empathy generation, which corresponds to the identification stage in Gross’s model, i.e., recognizing how the user feels.
Others elicit empathy through prompting and chain-of-thought reasoning~\citep{lin2024advancing}, which corresponds to the implementation stage, i.e., generating the empathetic response. However, they lack a principled, psychologically grounded framework for emotion regulation and address only a single stage in isolation. As a result, these systems merely react to the user's emotion rather than regulate it, producing responses that sound empathetic on the surface but are not tailored to what the user's emotional state actually calls for. Although text-based empathetic dialogue has been studied more extensively, through emotion-label conditioning, mimicry, and emotion-cause reasoning~\citep{rashkin2019empathetic,majumder2020mime,li2022kemp}, these methods share the same limitation, they map the perceived emotion directly to a response without deciding how it should be regulated.

Moreover, progress in empathetic spoken dialogue is limited by both data and evaluation. Spoken dialogue datasets contain natural speech but often non-empathetic or inconsistent responses~\citep{busso2008iemocap,li2018meld}, while empathetic dialogue datasets are text-only~\citep{rashkin2019empathetic,liu-etal-2021-towards}. Evaluation typically relies on text-similarity metrics~\citep{zhang2020bertscore}, emotion accuracy, or LLM judges~\citep{zheng2023judging}, which do not capture empathy and are rarely validated against human judgment.

To address this gap, we propose \textbf{ER-EDF}, a psychology-grounded framework for regulation-aware empathetic dialogue generation in large audio–language models (LALMs). ER-EDF structures LALM-based response generation around the three stages of Gross's Extended Process Model: it identifies the user's emotional state from speech, selects a regulation strategy matched to that state, and implements the strategy through the LALM. 
Moreover, we construct novel spoken empathetic dialogue datasets and introduce empathy-aware evaluation metrics. Evaluations across five LALMs and two datasets show that ER-EDF consistently induces more empathetic responses, validated by both objective metrics and human evaluation. Our contributions are summarized: 

\begin{itemize}
    \item We introduce \textbf{ER-EDF}, a model-agnostic framework that operationalizes Gross's 
    process model of emotion regulation for empathetic dialog generation 
    in LALMs. %
    
    \item %
    We construct an empathetic spoken dialogue 
    dataset built on IEMOCAP and MELD, providing supervision and evaluation resources for regulation-aware empathetic response generation. %
    
    \item We propose three empathy-aware evaluation metrics that go beyond lexical matching to capture empathetic behavior.%
\end{itemize}

%% file: latex/sections/related_work.tex
\section{Related Work}

\paragraph{Emotion Regulation and Gross model.}
In human conversation, empathic responses are themselves a form of interpersonal emotion regulation: listeners deliberately shape their partner's emotions by calming, validating, or sharing positive affect, and such regulation is central to supportive and empathic communication~\citep{zaki2013interpersonal,niven2009classification}. 
Emotion regulation theory studies how affective states are maintained,
intensified, reduced, or redirected over time~\citep{gross1998emerging}.%
Gross's Extended Process Model~\citep{Gross2015-bk} formalizes regulation as
a three-stage cycle: \emph{identification}, assessing whether and what
regulation is needed; \emph{selection}, choosing an appropriate strategy; and
\emph{implementation}, translating that strategy into a concrete action. Although originally formulated for regulating one's own emotions, this cycle extends to the interpersonal setting, where the regulator acts on another person's emotions~\citep{zaki2013interpersonal}. This implies that empathetic generation should not only recognize the user's emotion but also explicitly decide how the response should regulate it.

\paragraph{Empathetic Response Generation in LALMs.}
Recent LALM-based approaches for empathetic response generation broadly follow two directions. The first improves speech emotion recognition to guide empathetic response generation~\citep{xue2024echat, lin2024paralingpt, wang2024blspemo}, on the premise that a more accurate perception of the user’s emotion state leads to more empathetic responses.
In terms of Gross’s Extended Process Model~\citep{Gross2015-bk}, these methods strengthen the \emph{identification} stage. The second direction enhances response generation itself through prompting and chain-of-thought reasoning~\citep{lin2024advancing}, which corresponds to the \emph{implementation} stage, i.e., how a response is concretely realized.

However, neither direction incorporates a psychologically guided framework that elicits emotion regulation as humans do: empathy is expected to emerge implicitly from emotion-conditioned generation, rather than from a deliberate decision about whether and how the user's emotion should be regulated. ER-EDF addresses this gap by building such a framework, grounded in Gross's Extended Process Model, into LALM-based spoken dialogue.

\begin{figure*}[ht]
    \centering
    \includegraphics[width=\textwidth]{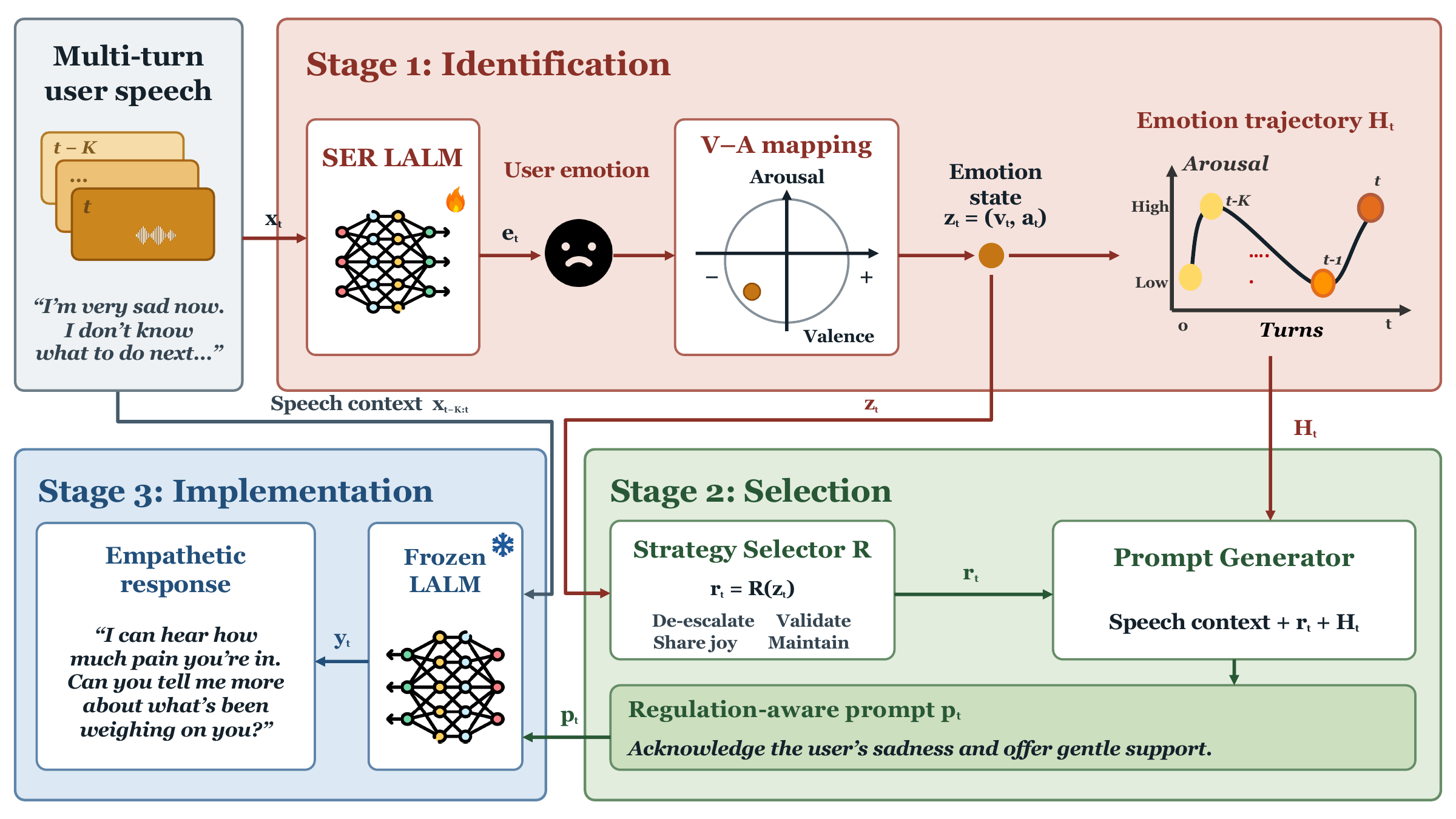}
    \caption{Overview of ER-EDF. The framework follows the three stages of Gross's Extended Process Model \citep{Gross2015-bk}: Identification (perceiving the emotional state), Selection (choosing a regulation strategy), and Implementation (generating the regulated response).}
    \label{fig:er-edf-framework}
\end{figure*}

\paragraph{Evaluation Metrics for Empathy.}
Standard %
evaluation metrics such as BLEU and ROUGE measure lexical overlap with reference 
responses, while embedding-based metrics such as BERTScore capture broader 
semantic similarity~\citep{zhang2020bertscore}. These are useful for fluency 
and reference matching, but they are not designed to assess whether a 
response expresses appropriate empathy. %
Recent work has turned to learned empathy classifiers, human preference judgments, and LLM-as-a-judge 
protocols~\citep{zheng2023judging}. While each of these is informative to a certain degree, the learned scores are difficult to interpret, human evaluation does not scale, and LLM judges introduce position, verbosity, and 
self-enhancement biases~\citep{zheng2023judging}. %
There is a need for %
evaluation metric that is empathy-sensitive, and grounded in the 
psychological constructs.%

%% file: latex/sections/method.tex
\section{Method}
\label{sec:method}

\subsection{Framework Overview}
\label{sec:method-overview}
Figure~\ref{fig:er-edf-framework} presents the overall architecture of
ER-EDF. The framework is explicitly organized around the three core stages
of Gross's Extended Process Model of emotion regulation
\citep{Gross2015-bk}: \emph{Identification}, which perceives and tracks
the user's emotional state from speech and assesses whether regulation is
warranted. Stage 2 of \emph{Selection} chooses a psychology-grounded
regulation strategy matched to the current affective state. Stage 3 of 
\emph{Implementation} finally generates the response based on the selected regulation strategy. %
Modeling all three stages %
allows ER-EDF to mirror how humans regulate others’ emotions in empathetic conversation ER-EDF first decides \emph{whether} to regulate, then
\emph{what} strategy to apply, and finally \emph{how} to realize that
strategy, using the Identification-Selection-Implementation cycle.

\subsubsection{Stage~1: Identification}
\label{sec:identification}

The Identification stage answers the first question posed by the Extended
Process Model: \emph{what is the user’s current emotional state, and does it warrant regulations?} ER-EDF operationalizes this stage in two steps:
speech emotion recognition and valence\-arousal mapping.

\paragraph{Speech emotion recognition.}
Given the current speech signal $\bm{x}_t$, a LALM fine-tuned for
speech emotion recognition predicts a categorical emotion label $e_t$
for the current turn. Finetuning details can be found in Appendix~\ref{app:settings}.

\paragraph{Valence-arousal mapping and trajectory tracking.}
\label{sec:identification}
According to Gross’s theory, the goal of emotion regulation is to change the occurrence, intensity, and direction of an emotional response~\citep{gross1998emerging}. Deciding whether and how to regulate therefore requires a representation in which these properties are explicit. Categorical emotion labels alone do not provide such a representation, as they neither encode the intensity of an emotion nor indicate the direction in which it should be regulated. The valence-arousal space~\citep{russell1980circumplex,reganger} directly captures these properties: valence indicates the direction of the affective state (positive or negative), and arousal indicates its intensity. ER-EDF therefore maps the predicted label $e_t$ onto a valence-arousal state $\bm{z}_t = (v_t, a_t)$ grounded in dimensional emotion theory (Appendix~\ref{app:map}). The resulting state $\bm{z}_t$ determines whether regulation is warranted and, if so, which regulation strategy is selected in Stage~2 (Section~\ref{sec:selection}). States that do not call for regulation (e.g., neutral) are passed on without an active regulation goal, which completes the identification decision of whether to regulate.

\subsubsection{Stage~2: Selection}
\label{sec:selection}

Given the identified emotional state $\bm{z}_t = (v_t, a_t)$, the %
Selection stage determines which regulation strategy to apply. %
We define a rule-based %
policy $r_t = R(\bm{z}_t)$ that maps each
valence-arousal region to a psychologically grounded interpersonal
regulation strategy~\citep{niven2009classification}:

\begin{table}[t]
\centering
\small
\resizebox{\columnwidth}{!}{%
\begin{tabular}{lccc}
\toprule
& \textbf{High} & \textbf{Medium} & \textbf{Low} \\
\midrule
\textbf{Negative}      & De-escalate      & Validate + support & Validate \\
\textbf{Positive}      & Share joy        & Maintain           & Maintain \\
\textbf{Mixed/Unknown} & Context-grounded & Context-grounded   & Context-grounded \\
\bottomrule
\end{tabular}%
}
\caption{Regulation strategy mapping from emotion states of valence and arousal. 
Full strategy descriptions and supporting references are provided in Appendix~\ref{app:strategies}.}%
\label{tab:strategy-core}
\end{table}
As shown in Table~\ref{tab:strategy-core}, for negative high-arousal states (e.g., anger, fear), de-escalation promotes calm and emotional containment
\citep{reganger}. For negative low-arousal states (e.g.,
sadness), validation offers acknowledgment without minimization
\citep{regsad1,regsad2}. Similarly, sharing joy amplifies positive experiences
\citep{regpos} and maintenance sustains neutral exchanges without
unnecessary intervention. For states with medium arousal or ambiguous
valence, the policy selects intermediate strategies that prioritize
the conversational content over a fixed regulation direction
(see Appendix~\ref{app:strategies} for the complete seven-strategy). The output of this stage is a strategy label $r_t$ that specifies the
\emph{regulatory goal} of the response: what it should accomplish
emotionally. %

\subsubsection{Stage~3: Implementation}
\label{sec:implementation}

The Implementation stage addresses the final question in Gross's model:
\emph{How is the selected strategy concretely realized?} 

ER-EDF assembles a structured prompt $p_t$ from three sources: (i) the preceding $K$ turns of the user's speech, which provide the conversational context; (ii) the emotion trajectory $H_t$(introduced in Stage ~\ref{sec:identification}, accumulated states over the recent $K$ turns), which expose how the user's affect has changed overtime and (iii) the selected strategy $r_t$, which specifies the regulatory goal of the reply (template in Appendix~\ref{sec:LLM prompts}). %
This prompt conditions a frozen LALM backbone to generate the final response $y_t$.
Supplying the trajectory and context alongside the strategy mirrors the monitoring and adjustment
loop described in the Extended Process Model \citep{Gross2015-bk}, where
regulation is not a one-shot decision but an ongoing process that adapts
to the evolving emotional state and further makes the response context-dependent: the same strategy is realized differently depending on what the user actually said, avoiding template-like empathetic replies.  %

\subsection{Dataset Construction}
\begin{figure*}[ht]{
    \centering
    \includegraphics[width=\textwidth]{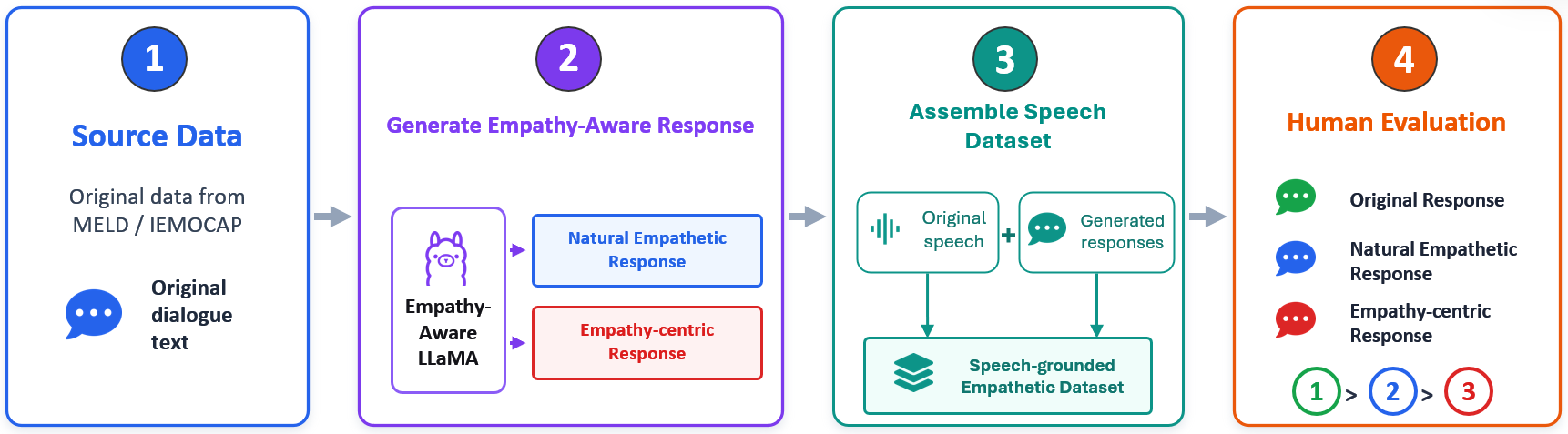}
    \caption{
    Dataset construction pipeline for empathy-aware response generation. }
    \label{fig:dataset-construction}
    }
\end{figure*}

Spoken conversation datasets provide natural speech and multi-turn dialogue context, but no empathetic response targets. We use two such dataset, IEMOCAP~\citep{busso2008iemocap} and MELD~\citep{li2018meld}. Their original speeches are frequently non-empathetic and sometimes even confrontational, and therefore cannot serve as empathetic targets. We thus generate an empathetic response for each turn as our dataset. As shown in Figure~\ref{fig:dataset-construction}, for each speech recording $\bm x_t$, we take the transcripts and use a text-based, empathy-focused model~\cite{empathyllama} to generate empathetic responses. %
The generated empathy-focused responses will place explicit emphasis on emotional support, validation, and empathic concern for the speaker. To ensure that the generated responses are empathetic and contextually appropriate, we include the preceding $K$ turns of transcriptions together as input for empathetic response generation. The full prompt templates for each type of response are provided in Appendix~\ref{sec:LLM prompts}. This is generated for two datasets, resulting in 2,610 single-turn empathetic dialogues for MELD and 5,666 for IEMOCAP (8,276 in total).

However, responses that foreground emotional support may become formulaic, expressing empathy mechanically without following the flow of the conversation. To guard against this, we additionally construct a second type, \emph{natural-empathetic responses}, which continue the dialogue coherently while expressing a proportionate level of empathy. They are generated with exactly the same pipeline and inputs as the empathy-focused responses, differing only in the generation prompt (Appendix~\ref{app:prompt-natural-empathetic}). Both generated responses are further validated by human annotators to verify their quality and validity in Section ~\ref{sec:Dataset-Validation}.

\subsection{Evaluation Metrics}
\paragraph{Empathy-Concern Index (ECI).}
We introduce the Empathy-Concern Index (ECI), a lexicon-based metric derived from the
World Well-Being Project (WWBP) empathy and distress lexica \citep{WWBP} which assign each word two scores: an empathy score, reflecting empathic concern and care directed at the other person (other-oriented); and a distress score, reflecting personal distress—the speaker's own discomfort or anxiety (self-oriented). This distinction follows Batson's account of empathy, in which empathic concern rather than personal distress drives supportive, other-oriented behavior~\citep{batson1991altruism}. Intuitively, ECI measures how much a response expresses concern for the listener relative to self-focused distress. Given a
generated response $y = (w_1, \dots, w_N)$, the base ECI is:
\begin{equation}
\resizebox{0.88\columnwidth}{!}{$\displaystyle
    S(y) = \frac{1}{N} \sum_{i=1}^{N}
    \mathbf{1}[w_i \in \mathcal{L}]
    \bigl(\tilde{\phi}_{\mathrm{emp}}(w_i) -
    \tilde{\phi}_{\mathrm{dist}}(w_i)\bigr)
$}
\end{equation}
where $\tilde{\phi}_{\mathrm{emp}}$ and $\tilde{\phi}_{\mathrm{dist}}$ are min-max
normalized empathy and distress weights from the lexicon $\mathcal{L}$, and $N$ is the
total token count (tokens absent from $\mathcal{L}$ contribute zero to the numerator
but are counted in $N$). 
A higher score indicates more other-oriented empathic concern
relative to distress-oriented language. We average over all $N$ words, not just those in $\mathcal{L}$, so that a response with only one empathic word does not score as high as one that is empathic throughout. 

To penalize degenerate outputs (e.g. repetitive or excessively long responses) and
reward contextual appropriateness, we apply an adjusted variant:
\begin{equation}
\label{eq:eci-full}
    \mathrm{ECI}(y) = S - |S|\,(1 - P_{\mathrm{rep}}\, P_{\mathrm{len}}\, P_{\mathrm{ctx}}),
\end{equation}
where $P_{\mathrm{rep}}$, $P_{\mathrm{len}}$, and $P_{\mathrm{ctx}} \in [0,1]$ penalize
lexical repetition, excessive length, and contextual mismatch respectively (definitions
in Appendix~\ref{app:eci-penalties}). When all penalties are inactive ($P=1$), the
formula reduces to the base score $S$.

\paragraph{Emo-BERT and Emp-BERT.}
We additionally introduce Emo-BERT and Emp-BERT, two empathy-aware variants of BERTScore~\citep{zhang2020bertscore}. Emo-BERT restricts matching to emotion-relevant tokens using the WWBP empathy and distress lexica, computing similarity only over filtered tokens. Emo-BERT thus measures alignment with emotion-focused content in the reference. Emp-BERT replaces the BERT backbone with EmoBERTa~\citep{kim2021emoberta}, enabling empathy-informed representations for scoring.

%% file: latex/sections/Experiments.tex
\section{Experimental Setup}
\label{sec:experimental-setup}

\paragraph{Datasets and settings.}The MELD~\cite{li2018meld} and IEMOCAP~\cite{busso2008iemocap} datasets are used for empathetic dialogue generation and evaluation.
For IEMOCAP, we report results averaged across five folds and for MELD, we report results on the official test split (2,610 utterances).

\paragraph{Compared systems.}

We evaluate ER-EDF across five audio-language model backbones: Audio Flamingo 3 \cite{goel2025audioflamingo3}, Gemma-4-E4B-IT \cite{gemmateam2026gemma4}, Megrez-3B-Omni \cite{li2025megrez}, MiDashengLM-7B \cite{dinkel2025midashenglm}, and Step-Audio 2 mini \cite{wu2025stepaudio2}. For each backbone, we compare ER-EDF against a \textbf{Baseline} that uses the same backbone with a generic empathetic-response prompt, i.e., without the proposed regulation principles. %
(Appendix~\ref{sec:LLM prompts}). All conditions share the same audio input, preceding-turn context, and decoding settings. We evaluate generation performance with context windows of $k \in \{1,3,5\}$. We additionally compare with existing open-source speech-based empathetic dialogue framework of BLSP-Emo \cite{BLSP-Emo} and OSUM-Echat \cite{OSUM-Echat}.

\paragraph{Dataset evaluation.}
We evaluate the quality of the constructed responses with both objective and subjective evaluations. The objective evaluation uses two types of reference-free metrics. The first type is Emo-BERT scoring applied in two settings: \emph{Resp-only Emp} scores the generated response on its own, measuring how empathetic it is as a standalone utterance, while \emph{Ctx+Resp Emp} scores the response together with its preceding five dialogue turns, measuring whether it remains empathetic within the conversation. We place particular weight on Emo-BERT, which shows the strongest and statistically significant alignment with human preferences (Section~\ref{sec:metric-eval}). The second type uses an LLM-as-a-judge~\citep{zheng2023judging} to rate each response for coherence with the dialogue context (\emph{LLM-Coh}) and naturalness as a conversational reply (\emph{LLM-Nat}); the full judging prompt is in Appendix~\ref{sec:LLM prompts}. 

For the subjective evaluation, we conduct a human study on a subset of 77 dialogue instances across the two datasets: 55 from IEMOCAP (11 per fold) and 22 from MELD. For each instance, annotators listen to the spoken dialogue context and rank three candidate responses, namely the original response, the natural-empathetic response, and the empathy-focused response, from most to least empathetic. 
For our empathy-aware dataset, a total of 9 reviewers participated in the evaluation and collected 549 human reviews in total. Each speech-dialogue instance received approximately 7.1 independent human judgments on average. This is measured by Win-Rate (the constructed dataset win rate agains original dataset) and Rank-1 rate (1st Rank rate for each dataset we constructed amoung all these three).

\paragraph{ER-EDF evaluation. } 
We finally evaluate the model responses via human evaluation and three proposed metrics (ECI, Emo-BERT, Emp-BERT), supplemented by BLEU/ROUGE (Appendix~\ref{app:Results-overlap}), between the baseline responses and ER-EDF responses.%
We report the human preference rate of ER-EDF against the baseline. Detailed settings are listed in Appendix~\ref{app:human} In total, we collect 240 human judgments over 180 evaluated instances with 6 reviewers. 

%% file: latex/sections/results.tex
\section{Results}

\subsection{Dataset Validation}%
\label{sec:Dataset-Validation}
\begin{table*}[ht]
\centering
\small
\setlength{\tabcolsep}{4pt}
\renewcommand{\arraystretch}{1.08}
\resizebox{\textwidth}{!}{%
\begin{tabular}{cccccccc}
\toprule
\textbf{Dataset} & \textbf{Reference}
& \multicolumn{4}{c}{\textbf{Objective Metrics}}
& \multicolumn{2}{c}{\textbf{Human Metrics}} \\
\cmidrule(lr){3-6} \cmidrule(lr){7-8}
& 
& \textbf{Resp-only Emp}
& \textbf{Ctx+Resp Emp}
& \textbf{LLM Coh}
& \textbf{LLM Nat}
& \textbf{Win Rate}
& \textbf{Rank-1 Rate} \\
\midrule
IEMOCAP & Original
& 0.6147 & 0.5707 & 4.1245 & 3.2030 & - & 7.5\%\\
IEMOCAP & Natural-empathetic
& 0.6277 & 0.5828 & \textbf{4.6598} & \textbf{4.2324} & 87.7\% & \textbf{50.9\%} \\
IEMOCAP & Empathy-focused
& \textbf{0.6548} & \textbf{0.5867} & 4.5837 & 4.1114 & \textbf{87.9\%} & 41.6\% \\
\midrule
MELD & Original
& 0.5602 & 0.5879 & 3.9720 & 4.0311 & - & 11.0\% \\
MELD & Natural-empathetic
& 0.6435 & 0.6100 & 4.4937 & 4.1348 & \textbf{85.3\%} & 37.7\% \\
MELD & Empathy-focused
& \textbf{0.6895} & \textbf{0.6383} & \textbf{4.7109} & \textbf{4.1500} & 83.8\% & \textbf{51.3\%} \\
\bottomrule
\end{tabular}
}
\caption{
Validation of the constructed empathetic responses against the original dialogue responses.}
\label{tab:reference-quality-validation}
\end{table*}

\begin{table*}[ht]
\centering
\small
\setlength{\tabcolsep}{3.5pt}
\renewcommand{\arraystretch}{1.1}
\resizebox{\textwidth}{!}{%
\begin{tabular}{llccccccccc}
\toprule
& & & \multicolumn{4}{c}{\textbf{MELD}} & \multicolumn{4}{c}{\textbf{IEMOCAP}} \\
\cmidrule(lr){4-7} \cmidrule(lr){8-11}
\textbf{Model} & \textbf{Setting} & \textbf{Human Overall}
& \textbf{ECI} & \textbf{EmoBERT} & \textbf{EmpBERT} & \textbf{Human}
& \textbf{ECI} & \textbf{EmoBERT} & \textbf{EmpBERT} & \textbf{Human} \\
\midrule
\multirow{2}{*}{Audio-Flamingo-3} & Baseline & 0.0\% & 0.0303 & 0.0003 & 0.7803 & 0.0\% & 0.0176 & -0.0224 & 0.7588 & 0.0\% \\
 & ER-EDF & \textbf{100.0\%} & \textbf{0.0367} & \textbf{0.0312} & \textbf{0.7917} & \textbf{100.0\%} & \textbf{0.0245} & \textbf{0.0122} & \textbf{0.7714} & \textbf{100.0\%} \\
\addlinespace
\multirow{2}{*}{Gemma-4-E4B-it} & Baseline & 21.7\% & 0.0232 & 0.0180 & 0.7989 & 16.7\% & 0.0242 & 0.0637 & 0.7959 & 27.3\% \\
 & ER-EDF & \textbf{78.3\%} & \textbf{0.0283} & \textbf{0.0404} & \textbf{0.8006} & \textbf{83.3\%} & \textbf{0.0347} & \textbf{0.1296} & \textbf{0.8080} & \textbf{72.7\%} \\
\addlinespace
\multirow{2}{*}{Megrez-3B-Omni} & Baseline & 30.4\% & 0.0358 & 0.0336 & 0.7851 & 33.3\% & 0.0349 & 0.0798 & 0.7897 & 27.3\% \\
 & ER-EDF & \textbf{69.6\%} & \textbf{0.0374} & \textbf{0.0522} & \textbf{0.7895} & \textbf{66.7\%} & \textbf{0.0376} & \textbf{0.1285} & \textbf{0.8015} & \textbf{72.7\%} \\
\addlinespace
\multirow{2}{*}{MiDashengLM-7B} & Baseline & 34.8\% & 0.0252 & -0.0001 & 0.7857 & 45.5\% & 0.0130 & 0.0278 & 0.7783 & 25.0\% \\
 & ER-EDF & \textbf{65.2\%} & \textbf{0.0265} & \textbf{0.0160} & \textbf{0.7894} & \textbf{54.5\%} & \textbf{0.0138} & \textbf{0.0688} & \textbf{0.7909} & \textbf{75.0\%} \\
\addlinespace
\multirow{2}{*}{Step-Audio-2-mini} & Baseline & 9.1\% & \textbf{0.0311} & -0.0194 & 0.7799 & 9.1\% & 0.0251 & -0.0020 & 0.7721 & 9.1\% \\
 & ER-EDF & \textbf{90.9\%} & 0.0287 & \textbf{0.0190} & \textbf{0.7874} & \textbf{90.9\%} & \textbf{0.0322} & \textbf{0.0662} & \textbf{0.7945} & \textbf{90.9\%} \\
\midrule
\multicolumn{11}{l}{\textit{other empathetic dialogue systems}} \\
BLSP-Emo$^{\dagger}$ & -- & -- & 0.0068 & -0.0227 & 0.7341 & -- & 0.0060 & 0.0801 & \textbf{0.8708} & -- \\
OSUM-Echat$^{\dagger}$ & -- & -- & 0.0098 & -0.0318 & 0.7282 & -- & 0.0127 & 0.0142 & 0.8120 & -- \\
\bottomrule
\end{tabular}%
}
\caption{ER-EDF response-generation performance on MELD and IEMOCAP.}
\label{tab:natural-results-main}
\end{table*} %
As shown in Table~\ref{tab:reference-quality-validation}, the constructed responses consistently improve upon the original responses, yielding higher scores on all objective empathy evaluation measures.
Human evaluation further confirms the quality of the constructed references. Both reference styles substantially outperform the original responses in ranking preference, achieving win rates of 83.8\%--87.9\% against the original responses. The Rank-1 results reveal a dataset-dependent preference pattern. Natural-empathetic references are favored on IEMOCAP, whereas empathy-focused references are preferred on MELD. We hypothesize that this difference stems from the distinct dialogue characteristics of the two datasets. IEMOCAP contains longer dyadic interactions in which emotions unfold over multiple turns, favoring responses that maintain conversational flow and express empathy more naturally. In contrast, MELD comprises short sitcom dialogues from Friends TV series, with rapid speaker exchanges, where explicit emotional acknowledgment is more salient~\cite{MELDisshorter,li2018meld}. Interestingly, the LLM-based evaluation aligns with human preferences, favoring natural-empathetic responses on IEMOCAP and empathy-focused responses on MELD. %
\subsection{Performance Comparison}
Table~\ref{tab:natural-results-main} reports the ER-EDF performance on five open-sourced model against each model baseline and two empathetic dialogue systems, under both \emph{natural-reference} and \emph{empathy-focused} reference for IEMOCAP and MELD. %

\noindent\textbf{Objective metrics. } ER-EDF achieves the best or comparable performance on all measures across both datasets. The gains in ECI indicate increased use of empathy-related lexical cues in generated responses, suggesting that ER-EDF encourages more explicit empathy-oriented language. Beyond lexical overlap, improvements on EmoBERT and EmpBERT further show enhanced semantic alignment in empathy-aware representation spaces.

Compared with prior speech empathetic dialogue framework, ER-EDF outperforms them across all metrics in meld, and show stronger eci and EmoBERT in IEMOCAP. %
Although BLSP-Emo achieves the highest EmpBERT score on IEMOCAP, ER-EDF variants obtain stronger ECI and EmoBERT scores on both datasets, suggesting that ER-EDF provides complementary gains in empathic lexicon coverage and empathic response quality. 

ER-EDF also improves lexical overlap with reference responses measured using BLEU and ROUGE metrics, across both datasets, ER-EDF outperforms baseline on nearly all BLEU and ROUGE metrics, with gains especially pronounced under the empathy-centric reference
(Appendix~\ref{app:Results-overlap}).
Overall, ER-EDF provides broad improvements across models and datasets.

\noindent\textbf{Human evaluation.} Specifically, ER-EDF shows consistently enhancement on human preference compared to the baselines. We further compute the average performance across five LALM backbones for each dataset. ER-EDF obtains overall average scores of 82.26\% on IEMOCAP and 79.08\% on MELD. 

We further examine whether ER-EDF's empathy gains come at the cost of
response diversity. As shown in Appendix~\ref{app:mattr-adaptivity},
ER-EDF responses show comparable or higher lexical diversity, measured by the moving-average type--token ratio (MATTR;~\citealp{covington2010mattr}), which computes the type-token ratio over a sliding 10-word window to remove length effects. They are also more adaptive to the speaker's emotion: $\eta^2$~\citep{richardson2011eta}, the proportion of variance in response sentence embeddings~\citep{reimers2019sentencebert} explained by the speaker's gold emotion, increases for all five models on both datasets.%

Overall, ER-EDF provides broad improvements across models, with gains varying depending on model capacity and capability.

\subsection{Metrics Evaluation}
\label{sec:metric-eval}
To ensure that the automatic metrics align with human judgment, that they reflect the empathy perceived by listeners rather than surface-level wording, we
collected pairwise preference annotations. For each pair, an annotator selected which response
was more empathetically appropriate. We then measured agreement between
human preferences and metric-implied rankings using three indicators:
raw agreement rate, point-biserial
correlation ($r_{\mathrm{pb}}$), and Kendall's $\tau_b$.

As shown in Table~\ref{tab:metric-agreement}, all three metrics achieve
agreement rates above .68 with human judgments, confirming their validity
as automatic proxies for empathetic quality. Both BERTScore-based metrics agree with human preferences in 75\% of the judgments, rising to 82\% on IEMOCAP. Moreover, the EmoBERT margin is significantly correlated with human preference ($\tau_b = 0.22$, $r_{\mathrm{pb}} = 0.25$, $p < 0.001$): the larger EmoBERT's advantage for a response, the more consistently humans prefer it.
\begin{table}[h]
\centering
\small
\resizebox{\columnwidth}{!}{%
\begin{tabular}{lccccc}
\toprule
 & \multicolumn{3}{c}{\textbf{Pairwise accuracy}} & \multicolumn{2}{c}{\textbf{Correlation}} \\
\cmidrule(lr){2-4}\cmidrule(lr){5-6}
\textbf{Metric} & \textbf{MELD} & \textbf{IEMOCAP} & \textbf{All} & \textbf{Kendall} $\boldsymbol{\tau_b}$ & $\boldsymbol{r_{\mathrm{pb}}}$ \\
\midrule
\textbf{EmoBERT} & \textbf{0.693} & \textbf{0.818} & \textbf{0.755} & \textbf{0.223}$^{***}$ & \textbf{0.254}$^{***}$ \\
EmpBERT          & 0.683          & \textbf{0.818} & 0.750          & 0.131$^{*}$            & 0.105 \\
ECI              & 0.594          & 0.778          & 0.685          & 0.035                  & 0.040 \\
\bottomrule
\end{tabular}%
}
\caption{Alignment between automatic metrics and human pairwise preferences, $^{***}p<0.001$ (two-sided significance test of the correlation).}
\label{tab:metric-agreement}
\end{table}
We place particular weight on Emo-BERT, which shows the strongest and statistically significant alignment with human preferences (Section~\ref{sec:metric-eval}).
\begin{table*}[t]
\centering
\fontsize{8pt}{9.5pt}\selectfont
\setlength{\tabcolsep}{5pt}
\renewcommand{\arraystretch}{1.1}
\begin{tabular}{lccccccccc}
\toprule
& \multicolumn{3}{c}{\textbf{Components}}
& \multicolumn{3}{c}{\textbf{IEMOCAP}}
& \multicolumn{3}{c}{\textbf{MELD}} \\
\cmidrule(lr){2-4} \cmidrule(lr){5-7} \cmidrule(lr){8-10}
\textbf{Variant}
& \textbf{Emo.} & \textbf{Reg.} & \textbf{SER}
& \textbf{ECI} & \textbf{EmoBERT} & \textbf{EmpBERT}
& \textbf{ECI} & \textbf{EmoBERT} & \textbf{EmpBERT} \\
& \textbf{hint} & \textbf{rules} & \textbf{adapter}
& & & & & & \\
\midrule
SER-tuned Backbone
& $\times$ & $\times$ & \checkmark
& 0.0214 & $-$0.0432 & 0.7449
& 0.0283 & $-$0.0111 & 0.7810 \\
Emotion w/o regulation
& \checkmark & $\times$ & $\times$
& 0.0174 & $-$0.0225 & 0.7570
& 0.0321 & $-$0.0005 & 0.7797 \\
Regulation w/o emotion
& $\times$ & \checkmark & $\times$
& 0.0215 & 0.0011 & 0.7686
& 0.0334 & 0.0254 & 0.7916 \\
\midrule
Full ER-EDF
& \checkmark & \checkmark & $\times$
& \textbf{0.0245} & \textbf{0.0122} & \textbf{0.7714}
& \textbf{0.0367} & \underline{0.0312} & \underline{0.7917} \\
\midrule
Oracle ER-EDF
& GT & \checkmark & $\times$
& \underline{0.0242} & \underline{0.0070} & \underline{0.7692}
& \underline{0.0361} & \textbf{0.0314} & \textbf{0.7922} \\
\bottomrule
\end{tabular}
\caption{Ablation study of ER-EDF on Audio Flamingo 3. \emph{Emo.\ hint}: SER-predicted emotion label and valence-arousal state in the stage 1 Identification(GT, ground-truth labels). \emph{Reg.\ rules}: regulation instructions in the stage 2 (Selection). \emph{SER adapter}: the generator itself is the SER-fine-tuned LALM. Oracle ER-EDF serves as an upper-bound reference. Bold and underlined values indicate the best and second-best scores in each column, respectively.}
\label{tab:component-ablation}
\end{table*}

\subsection{Ablation Study}
\label{sec:ablation}
We ablate the components of ER-EDF to disentangle their contributions (Table~\ref{tab:component-ablation}). Each variant is defined by what it removes from Full ER-EDF. \textbf{Emotion w/o regulation} removes the regulation rules but keeps the emotion from stage 1. \textbf{Regulation w/o emotion} keeps the rules but removes all emotion hints. \textbf{SER-tuned Backbone} removes both, and instead uses the SER-fine-tuned LALM as the generator, testing whether stronger perception inside the generator alone yields more empathetic responses. Finally, \textbf{Oracle ER-EDF} replaces predicted emotion labels with ground-truth labels, providing an upper-bound reference for perfect identification.

The results lead to three observations. First, regulation contributes more than perception alone: Regulation w/o emotion outperforms both perception-only variants on all three metrics across both datasets (e.g., EmoBERT 0.0011 vs.\ $-$0.0225 and $-$0.0432 on IEMOCAP), whereas Emotion w/o regulation performs on par with the baseline (Table~\ref{tab:natural-results-main}), showing that supplying emotion labels without a regulation strategy brings little benefit. Second, both stages are needed: Full ER-EDF outperforms Regulation w/o emotion on all metrics, indicating that explicitly identifying the user's emotion makes the selected regulation more effective than letting the LALM infer it implicitly. Third, Full ER-EDF performs on par with or slightly above Oracle ER-EDF
(differences below 0.006 on all metrics). On MELD, where emotion labels
are derived from naturally scripted dialogue, Oracle ER-EDF marginally
outperforms Full ER-EDF, confirming that better perception helps. On
IEMOCAP, however, Full ER-EDF slightly exceeds the oracle variant,
likely because IEMOCAP's acted elicitation protocol produces more
extreme and less naturalistic emotion labels, so ground-truth labels
can over-specify the affective state and lead to regulation that is
disproportionate to the actual conversational context.

%% file: latex/sections/conclusion.tex
\section{Conclusion}

In this work, we presented ER-EDF, a framework that builds LALM-based empathetic dialogue around the three stages of human emotion regulation described by Gross's Extended Process Model. 
ER-EDF identifies the user's emotion with an SER-fine-tuned LALM and maps it to a valence–arousal state, selects a psychologically grounded regulation strategy for that state, and implements the strategy by conditioning a frozen LALM on the strategy, the user's emotional states, and the dialogue context, all three stages are aligned with 3 stages of Gross's Extended Process Model.
Across five LALMs and two datasets, ER-EDF improves empathy-aware metrics and human preference, and ablations show that regulation contributes beyond emotion perception alone. These results suggest that following how humans regulate others' emotions, rather than reacting to perceived emotion directly, is a promising path toward empathetic spoken dialogue systems.

%% file: latex/appendix/prompts.tex
\subsection{LLM prompts}
\label{sec:LLM prompts}
This appendix documents the prompt templates used for the current response-generation experiments and component ablations. It also records the dataset-specific templates associated with the reused natural empathetic and empathy-focused references. Placeholders denote sample-specific inputs and are not literal model instructions. Reference replies are used only for evaluation and are never included in response-generation prompts.

\paragraph{Baseline Prompt}
The baseline uses the original backbone without explicit SER hints. Up to three preceding audio clips are followed by the current utterance. If fewer preceding turns are available, only those turns are included.

\textbf{System prompt.}

\begin{PromptBox}
You are the other speaker in this conversation. Listen to the audio and reply naturally and empathetically to the latest speaker. Stay grounded in what was actually said. Output only the next spoken reply in one or two short sentences. Do not narrate, analyze, mention emotion labels, or continue the latest speaker's sentence.
\end{PromptBox}

\textbf{User prompt.}

\begin{PromptBox}
Previous conversation:
<audio>
<audio>
<audio>
Latest speaker:
<audio>
Give the next reply.
\end{PromptBox}

Each audio placeholder is replaced by the corresponding audio input through the backbone's native interface; the preceding-turn placeholders are omitted when unavailable.

\paragraph{Emotion Regulation Prompt with SER Predictions}
\label{app:prompt-emoreg-ser}

The full ER-EDF prompt consists of three components:
audio dialogue context, an SER-derived emotion hint,
and fixed conditional regulation instructions.
These components are assembled into a system message
and a user message, as described below.

\textbf{Audio dialogue context.}
The user message presents the available preceding audio
turns in chronological order, followed by the latest
speaker's audio. The audio placeholders below represent
model-specific audio inputs rather than literal prompt text.

\begin{PromptBox}
Previous conversation:
[audio context: previous turns in chronological order]
Latest speaker:
[current audio]
Give the next reply.
\end{PromptBox}

\textbf{SER-derived emotion hint.}
The predicted dataset-specific emotion label is mapped
to a coarse valence--arousal state using the fixed
lookup table in Appendix~\ref{app:map}.
The following hint is appended to the user message:

\begin{PromptBox}
Speech emotion hint: [emotion label]. State: [valence]/[arousal].
\end{PromptBox}

When provided in the input manifest, the previous state
of the same speaker is appended to the same hint line:

\begin{PromptBox}
Previous state of this same speaker: [previous valence]/[previous arousal].
\end{PromptBox}

This optional field supplies a previous-state cue.
The explicit emotion information therefore consists of
the current label and state, together with the previous
same-speaker state when available.

Valence takes the values negative, neutral, positive,
or mixed; arousal takes the values low, medium, or high.
Disgust maps to negative/medium, and surprise maps to
mixed/high. Missing or unrecognized labels are rendered
as ``unavailable'', with state ``unknown/unknown''.

\textbf{State-conditioned regulation strategy.}
Before generation, the rule-based policy $R(\bm{z}_t)$ selects one
strategy $r_t$ for each sample from the seven state-specific strategies
in Table~\ref{tab:state-strategy-map}, according to the identified
valence--arousal state; missing or unrecognized labels are mapped to the
\texttt{unknown/unknown} strategy. The description of the selected
strategy is appended to the baseline system prompt:

\begin{PromptBox}
Regulation strategy: [selected strategy name]. [selected strategy description]
\end{PromptBox}

The instruction therefore varies across samples only through the
selected strategy. Negative and positive high-arousal states receive
explicit regulation strategies (e.g., de-escalation, validation, or
sharing joy), whereas mixed, neutral, and unknown states receive
goal-based guidance that keeps the reply grounded in the conversational
content without imposing a regulation direction. The strategy descriptions
elaborate the intended response behavior and do not prescribe fixed
templates. Because $r_t$ is selected by the external policy before
generation and explicitly included in the prompt, the LALM performs
strategy implementation rather than strategy selection.

\textbf{Prompt assembly and response generation.}
The complete system message combines the baseline
response instructions with the conditional regulation
instructions:

\begin{PromptBox}
You are the other speaker in this conversation. Listen to the audio and reply naturally and empathetically to the latest speaker. Stay grounded in what was actually said. Output only the next spoken reply in one or two short sentences. Do not narrate, analyze, mention emotion labels, or continue the latest speaker's sentence. For negative high arousal, de-escalate; for negative low arousal, validate and support; for positive high arousal, share joy; otherwise maintain a natural conversation.
\end{PromptBox}

The complete user message consists of the audio-context
component followed by the SER-derived emotion hint,
including the optional previous-state sentence when
available. The generator produces only the next spoken
reply.

\textbf{SER prediction source.}
In the reported main experiments, Audio Flamingo 3,
Gemma-4-E4B-IT, MiDashengLM-7B, and Step-Audio 2 mini
use frozen predictions from the fine-tuned
Qwen2.5-Omni-7B SER model.
Megrez-3B-Omni uses predictions from its own
fine-tuned SER model.
\paragraph{Ablation Prompts}
All component ablations preserve the same audio context and decoding configuration. Shared prompt text is reused exactly; the changes below specify each ablation completely.

\textbf{Emotion without regulation.}
This setting uses the baseline system prompt and the full ER-EDF user prompt, including the SER emotion and state hints. Thus, only the regulation-principle sentence is removed; no additional instruction prohibiting regulation is inserted.

\textbf{Regulation without emotion hints.}
This setting uses the full ER-EDF system prompt and the baseline user prompt. The regulation principles remain unchanged, while the current emotion label, current state, and optional previous-speaker state are all removed. This is also the matched-rules control used in development experiments.

\textbf{Oracle ER-EDF.}
This setting uses the same system and user templates as full ER-EDF, replacing the predicted emotion labels with the corresponding ground-truth labels before state mapping. The availability of the previous-state field remains unchanged. The prompt still says ``Speech emotion hint''; no oracle-specific instruction is added. Ground-truth emotion labels are supplied only in this diagnostic condition.

\textbf{SER-tuned backbone.}
This setting uses exactly the baseline system and user prompts, with the generator's own SER adapter loaded. No emotion labels, state summaries, or regulation principles are inserted. MELD uses its dataset-specific adapter, and IEMOCAP uses the adapter trained without the test session.

\paragraph{Natural Empathetic Reference Prompt}
\label{app:prompt-natural-empathetic}
The reused reference pools were prepared with dataset-specific templates. The natural empathetic setting emphasizes conversational coherence and contextually appropriate empathy.

\textbf{MELD system prompt.}

\begin{PromptBox}
You are writing a high-quality reference reply for a two-person conversation.
Primary goal: continue the dialogue naturally and fluently while showing clear empathy.
Keep the flow coherent with the latest utterance and prior context.

Core requirements:
1) Continue the conversation naturally: respond to what was just said and add one small next step.
2) Empathy should be present but integrated naturally (not preachy or overly formal).
3) Keep it concise, conversational, and scene-consistent.

Hard constraints:
- Output ONLY the next reply text.
- No analysis, no bullet points, no narration/stage directions.
- No quotes wrapping the whole reply.
\end{PromptBox}

\textbf{MELD user prompt.} The transcript includes the preceding turns and the current utterance, with speaker identifiers.

\begin{PromptBox}
Conversation:
[dialogue transcript]

Write the next reply:
\end{PromptBox}

\textbf{IEMOCAP system prompt.}

\begin{PromptBox}
You write a high-quality reference reply for a two-person spoken dialogue.
Primary goal: continue the conversation naturally and fluently while keeping clear empathy.
Be coherent with the latest utterance and dialogue context.
Output only one concise spoken reply (1-2 sentences). No analysis.
\end{PromptBox}

\textbf{IEMOCAP user prompt.}

\begin{PromptBox}
This is an ongoing conversation. Here is the context:

[dialogue transcript]

Latest speaker: Speaker [current speaker]
Latest utterance: "[current utterance]"

Write the next reply that sounds natural, conversational, and empathetic. Acknowledge feelings briefly and keep the dialogue moving with one small next step.
Output only the reply text.
\end{PromptBox}

When no preceding context is available, the IEMOCAP user prompt becomes:

\begin{PromptBox}
Someone just said: "[current utterance]"

Write the next reply that sounds natural, conversational, and empathetic. Acknowledge feelings briefly and keep the dialogue moving with one small next step.
Output only the reply text.
\end{PromptBox}

\paragraph{Empathy-Focused Reference Prompt}
\label{app:prompt-empathy-centric}
The empathy-focused references used in the current evaluation correspond to \texttt{empathetic\_response\_focus}. They emphasize explicit acknowledgment, validation, and a supportive next step. This reference pool is distinct from the separately implemented \texttt{empathy\_centric\_dialogue} style.

\textbf{MELD system prompt.}

\begin{PromptBox}
You are writing a high-quality reference reply for a two-person conversation.
Primary goal: produce a strongly empathetic response using warm, validating language and attitude.
The reply should still fit the conversation and move it forward.

Core requirements:
1) Explicitly acknowledge the speaker's emotion or situation.
2) Use empathetic wording (validation, support, understanding) in natural spoken English.
3) Add one helpful next step (gentle question, suggestion, or supportive follow-up).

Hard constraints:
- Output ONLY the next reply text.
- No analysis, no bullet points, no narration/stage directions.
- No quotes wrapping the whole reply.
\end{PromptBox}

\textbf{MELD user prompt.} The same transcript-based user template is used as for the natural empathetic references.

\begin{PromptBox}
Conversation:
[dialogue transcript]

Write the next reply:
\end{PromptBox}

\textbf{IEMOCAP system prompt.}

\begin{PromptBox}
You write a high-quality reference reply for a two-person spoken dialogue.
Primary goal: produce a strongly empathetic response using warm, validating language.
Keep it coherent with context and still move the dialogue forward.
Output only one concise spoken reply (1-2 sentences). No analysis.
\end{PromptBox}

\textbf{IEMOCAP user prompt.}

\begin{PromptBox}
This is an ongoing conversation. Here is the context:

[dialogue transcript]

Latest speaker: Speaker [current speaker]
Latest utterance: "[current utterance]"

Write the next reply with explicit empathy and validation, then add a gentle supportive follow-up (a question, suggestion, or reassurance). Keep it natural and concise.
Output only the reply text.
\end{PromptBox}

When no preceding context is available, the IEMOCAP user prompt becomes:

\begin{PromptBox}
Someone just said: "[current utterance]"

Write the next reply with explicit empathy and validation, then add a gentle supportive follow-up (a question, suggestion, or reassurance). Keep it natural and concise.
Output only the reply text.
\end{PromptBox}

The current evaluation averages scores against the natural empathetic and empathy-focused references with equal weights within each dataset. The response-generation and ablation runs reuse these reference pools; they do not generate new references.

%% file: latex/appendix/method_app.tex
\subsection{Valence-Arousal Mapping Rules}
\label{app:map}
We map each discrete emotion label to a coarse valence--arousal
state using a fixed lookup table. Valence takes one of four values:
negative, neutral, positive, or mixed; arousal takes one of three
values: low, medium, or high. These states are label-derived proxies
used to condition response generation, rather than continuous
valence--arousal measurements. Missing or unrecognized labels are
assigned \texttt{unknown/unknown}.

\begin{table}[htbp]
\centering
\small
\renewcommand{\arraystretch}{1.08}
\caption{Valence-arousal mapping rules implemented for MELD emotion labels.}
\label{tab:meld-va-map}
\begin{tabular}{lll}
\toprule
\textbf{MELD Label} & \textbf{Valence} & \textbf{Arousal} \\
\midrule
Anger    & Negative & High \\
Fear     & Negative & High \\
Disgust  & Negative & Medium \\
Sadness  & Negative & Low \\
Joy      & Positive & High \\
Surprise & Mixed    & High \\
Neutral  & Neutral  & Low \\
\bottomrule
\end{tabular}
\end{table}

\begin{table}[htbp]
\centering
\small
\renewcommand{\arraystretch}{1.08}
\caption{Valence--arousal mapping rules implemented for IEMOCAP emotion labels.}
\label{tab:iemocap-va-map}
\begin{tabular}{lll}
\toprule
\textbf{IEMOCAP Label} & \textbf{Valence} & \textbf{Arousal} \\
\midrule
Angry   & Negative & High \\
Sad     & Negative & Low \\
Happy   & Positive & High \\
Neutral & Neutral  & Low \\
\bottomrule
\end{tabular}
\end{table}

\subsection{Emotion Regulation Strategies}
\label{app:strategies}

\begin{table*}[htbp]
\centering
\small
\setlength{\tabcolsep}{6pt}
\renewcommand{\arraystretch}{1.18}
\begin{tabular}{
p{0.10\textwidth}
p{0.09\textwidth}
p{0.70\textwidth}
}
\toprule
\textbf{Valence} & \textbf{Arousal} & \textbf{Strategy} \\
\midrule

Negative & High
& \textbf{De-escalation and grounding.}
Acknowledge the concrete concern expressed by the speaker and
respond in a calm, measured manner. Use clear, concise language
that addresses the immediate situation without intensifying
conflict or mirroring hostile wording. Keep the reply grounded
in what the speaker actually said, and, when appropriate,
offer a manageable next step that helps the conversation proceed
with less tension. \\

\addlinespace
Negative & Medium
& \textbf{Validation and context-sensitive support.}
Recognize the difficulty or discomfort conveyed by the speaker
without exaggerating its intensity or imposing an emotional
interpretation. Briefly validate the concern and offer a gentle,
relevant next step when the context supports one. The response
should remain specific to the situation, avoiding generic
reassurance, unsolicited advice, or pressure to adopt a more
positive attitude. \\

\addlinespace
Negative & Low
& \textbf{Validation and gentle encouragement.}
Acknowledge the speaker's experience with a supportive,
non-judgmental response. Allow space for the concern rather than
immediately attempting to resolve it or redirecting the
conversation toward a positive interpretation. When appropriate,
add a small offer of support, a gentle question, or an encouraging
next step that follows naturally from the dialogue. Avoid
dismissing the concern or making unsupported promises. \\

\addlinespace
Positive & High
& \textbf{Sharing positive emotion.}
Respond warmly to the positive experience and share the speaker's
enthusiasm while maintaining conversational coherence.
Connect the reply to the specific event or detail mentioned,
rather than offering a generic expression of excitement.
Keep the level of enthusiasm balanced and appropriate to the
scene, and continue the exchange with a relevant observation,
question, or next step without introducing unnecessary concern
or emotional correction. \\

\addlinespace
Mixed & High
& \textbf{Content-focused response with cautious clarification.}
Avoid assuming that the speaker's reaction is exclusively positive
or negative. Respond first to the event, question, or concern that
is explicitly expressed, using language that remains compatible
with more than one emotional interpretation. If the ambiguity
affects how to continue the conversation, ask a brief,
non-leading clarification question. Otherwise, continue naturally
without forcing reassurance, celebration, or an explicit
description of the speaker's feelings. \\

\addlinespace
Neutral & Low
& \textbf{Natural conversational continuation.}
Respond directly to the communicative purpose of the utterance,
such as answering a question, acknowledging information, or
continuing an ongoing exchange. Maintain a natural, empathetic
tone while keeping the reply proportionate to the context.
Do not introduce emotional validation, reassurance, or celebration
solely because an emotion-regulation framework is being used.
Let the conversational content determine the appropriate next
response. \\

\addlinespace
Unknown & Unknown
& \textbf{Uncertainty-aware, context-grounded response.}
Treat missing or unrecognized emotion labels as insufficient
evidence for a specific emotional interpretation. Rely on the
available audio dialogue context and respond to what can be
understood from the speaker's utterance. Avoid diagnosing a
feeling or forcing a particular regulation direction. Ask for
clarification only when the uncertainty prevents an appropriate
response; otherwise, provide a concise and natural continuation
grounded in the available information. \\

\bottomrule
\end{tabular}
\caption{State-to-strategy mapping, prioritizing explicit regulation
rules and using goal-based guidance for states not covered by those
rules. The descriptions elaborate the intended response behavior;
they do not prescribe fixed response templates.}
\label{tab:state-strategy-map}
\end{table*}
\section{ECI Penalty Definitions}
\label{app:eci-penalties}

The corrected ECI (Eq.~\ref{eq:eci-full}) applies three multiplicative penalty factors.
Let $U$ be the number of distinct lexicon-matched tokens, $M$ the total number of
lexicon-matched token occurrences, $n$ the response length in words, and
$C = \frac{1}{N}\sum_{i}\mathbf{1}[w_i \in \mathcal{L}]\,\tilde{\phi}_{\mathrm{emp}}(w_i)$
the mean empathy weight of the response.

\paragraph{Repetition penalty.}
\begin{equation}
    P_{\mathrm{rep}} = \frac{U}{M}.
\end{equation}
$P_{\mathrm{rep}} = 1$ when every matched token is unique; it decreases toward zero as
repetition increases. If no tokens match the lexicon, $P_{\mathrm{rep}} = 0$.

\paragraph{Length penalty.}
\begin{equation}
    P_{\mathrm{len}} = \exp\!\left[-\,\frac{\max\bigl(0,\; n - (12 + 24\,E^{\star})\bigr)}{12}\right],
\end{equation}
which equals 1 when the response is at or below a target length that scales with the
expected empathy intensity $E^{\star}$, and decays exponentially beyond it.

\paragraph{Context penalty.}
\begin{equation}
    P_{\mathrm{ctx}} = \exp\!\left[-(6 - 4\,E^{\star})\,\max(0,\; C - E^{\star})\right],
\end{equation}
which penalizes responses whose empathy density $C$ exceeds the contextually expected
level $E^{\star}$.

\paragraph{Expected empathy intensity.}
$E^{\star}$ is derived from the gold-standard emotional state of the dialogue turn:
\begin{equation}
    E^{\star} = \mathrm{clip}\!\left(0.10 + 0.35\,(0.65\,V + 0.35\,A),\; 0.12,\; 0.55\right),
\end{equation}
where $V$ and $A$ are ordinal encodings of valence and arousal
($V_{\text{neg, mix, neu, pos}} = \{1, 0.8, 0.45, 0.3\}$;
$A_{\text{high, med, low}} = \{1, 0.7, 0.4\}$). All coefficients are manually specified
hyperparameters.

%% file: latex/appendix/SER.tex
\subsection{SER performance}
\label{app:SER}
\begin{table}[htbp]
\centering
\small
\label{tab:ser_finetune_dataset}
\begin{tabular}{llrr}
\toprule
\textbf{Model} & \textbf{Dataset} & \textbf{Baseline} & \textbf{Finetuned} \\
\midrule
Audio Flamingo 3 & MELD & 45.86 & \textbf{55.94} \\
Audio Flamingo 3 & IEMOCAP & 35.15 & \textbf{65.86} \\
\midrule
Gemma-4-E4B-IT & MELD & 45.13 & \textbf{51.07} \\
Gemma-4-E4B-IT & IEMOCAP & 27.10 & \textbf{60.74} \\
\midrule
Megrez-3B-Omni & MELD & 26.05 & \textbf{53.60} \\
Megrez-3B-Omni & IEMOCAP & 0.65 & \textbf{62.97} \\
\midrule
MiDashengLM-7B & MELD & 52.49 & \textbf{53.75} \\
MiDashengLM-7B & IEMOCAP & 32.87 & \textbf{43.27} \\
\midrule
Step-Audio 2 mini & MELD & \textbf{55.48} & 54.37 \\
Step-Audio 2 mini & IEMOCAP & 28.74 & \textbf{63.02} \\
\bottomrule
\end{tabular}
\caption{Speech emotion recognition accuracy before and after SER fine-tuning on MELD and IEMOCAP.}
\end{table}

%% file: latex/appendix/experiment_settings.tex
\subsection{Experiment setting}
\label{app:settings}

\paragraph{SER fine-tuning protocol.}
SER is formulated as audio-conditioned generation of an emotion label.
MELD uses seven labels and its training, validation, and test splits,
containing 9,988, 1,108, and 2,610 utterances, respectively.
IEMOCAP uses the project's four-class label mapping and five
session-held-out folds; each adapter is trained on four sessions
and evaluated on the remaining session. The pooled SER test set
contains 5,665 utterances.
The SER runs use LoRA with rank $r=8$, $\alpha=32$, and dropout $0.05$.
The learning rate is $10^{-4}$, the maximum gradient norm is $1.0$,
and the random seed is 42. The nominal effective batch size is
64 utterances. The archived and newly completed runs use the
distinct optimization settings specified below.

\paragraph{Archived SER training settings.}
The archived Megrez, MiDash, and Qwen2.5-Omni-7B runs on both
datasets, together with Step on MELD, were configured for five
epochs using AdamW with $(\beta_1,\beta_2)=(0.9,0.95)$,
$\epsilon=10^{-8}$, weight decay $0.1$, cosine learning-rate decay,
and a warmup ratio of $0.05$.
They use BF16, a per-device batch size of 4,
16 gradient-accumulation steps, and a maximum sequence length
of 2,048 tokens.
Megrez on MELD instead uses batch size 1, 64 accumulation steps,
and a maximum sequence length of 1,024 tokens.
LoRA targets the eligible linear layers through the ms-swift
\texttt{all-linear} configuration, subject to the model-specific
encoder and projector freezing settings.
Gradient checkpointing is enabled for Megrez, Qwen, and Step,
and disabled for MiDash.
We reuse the archived adapters; Step's MELD SER evaluation uses
the latest available adapter at optimization step 600.

\paragraph{Additional SER training settings.}
Audio-Flamingo-3 and Gemma-4-E4B-it on both datasets, and
Step-Audio-2-mini on IEMOCAP, are trained for five epochs using
AdamW with $(\beta_1,\beta_2)=(0.9,0.999)$, $\epsilon=10^{-8}$,
and weight decay $0.01$.
These runs use a constant learning rate of $10^{-4}$ without
warmup or a scheduler, batch size 1, and 64 gradient-accumulation
steps.
LoRA is applied to the language model's attention projections
(\texttt{q\_proj}, \texttt{k\_proj}, \texttt{v\_proj}, and
\texttt{o\_proj}) and feed-forward projections
(\texttt{gate\_proj}, \texttt{up\_proj}, and \texttt{down\_proj}),
with no trainable biases.
Base-model parameters, including acoustic encoders, remain frozen.
The base weights use BF16, and non-reentrant gradient
checkpointing is enabled.

\paragraph{Response-generation hyperparameters.}
Each input contains the current audio utterance and up to three
preceding turns ($k=1, 3, 5$), ordered chronologically.
Both Baseline and ER-EDF generate one response per input using
BF16 weights, batch size 1, temperature $0.5$, nucleus sampling
with $\texttt{top\_p}=0.9$, and at most 96 new tokens.
The random seed is reset to 42 before each request.
Audio-Flamingo-3 and Gemma explicitly disable top-$k$ filtering
($\texttt{top\_k}=0$); Megrez, MiDash, and Step retain the
generation-stack default ($\texttt{top\_k}=50$).
Other decoding options retain their model and runtime defaults.
Prompts request one or two short spoken sentences.
Megrez additionally receives ``Answer in at most 15 words.''
in both the Baseline and ER-EDF system prompts; this is an
instruction rather than a hard post-generation word limit.
Within each backbone, the two conditions use identical audio
context and decoding settings.
The AF and Gemma component ablations retain these same
generation settings.

%% file: latex/appendix/results_app.tex
\subsection{Extra Results}
\paragraph{Results on Lexicon-Overlap Metrics}
\label{app:Results-overlap}
\begin{table*}[htbp]
\centering
\scriptsize
\label{tab:appendix_text_metrics_two_refs}
\resizebox{\textwidth}{!}{%
\begin{tabular}{llrrrrrrr}
\toprule
\textbf{Model} & \textbf{Setting} & \textbf{B-1} & \textbf{B-2} & \textbf{B-3} & \textbf{B-4} & \textbf{R-1} & \textbf{R-2} & \textbf{R-L} \\
\midrule
\multicolumn{9}{l}{\textit{Natural empathetic reference -- IEMOCAP}} \\
Audio Flamingo 3 & Baseline & 0.049 & 0.015 & 0.008 & 0.006 & 0.141 & 0.022 & 0.113 \\
Audio Flamingo 3 & ER-EDF & \textbf{0.057} & \textbf{0.018} & \textbf{0.010} & \textbf{0.007} & \textbf{0.159} & \textbf{0.027} & \textbf{0.128} \\
Gemma-4-E4B-IT & Baseline & 0.059 & 0.021 & 0.012 & 0.009 & 0.184 & 0.036 & 0.147 \\
Gemma-4-E4B-IT & ER-EDF & \textbf{0.064} & \textbf{0.024} & \textbf{0.014} & \textbf{0.010} & \textbf{0.203} & \textbf{0.044} & \textbf{0.162} \\
Megrez-3B-Omni & Baseline & 0.083 & 0.025 & 0.014 & 0.010 & 0.195 & 0.036 & 0.154 \\
Megrez-3B-Omni & ER-EDF & \textbf{0.107} & \textbf{0.033} & \textbf{0.018} & \textbf{0.012} & \textbf{0.228} & \textbf{0.046} & \textbf{0.174} \\
MiDashengLM-7B & Baseline & 0.042 & 0.014 & 0.008 & 0.006 & 0.152 & 0.026 & 0.128 \\
MiDashengLM-7B & ER-EDF & \textbf{0.056} & \textbf{0.019} & \textbf{0.011} & \textbf{0.008} & \textbf{0.175} & \textbf{0.031} & \textbf{0.145} \\
Step-Audio 2 mini & Baseline & 0.057 & 0.019 & 0.011 & 0.007 & 0.160 & 0.031 & 0.132 \\
Step-Audio 2 mini & ER-EDF & \textbf{0.100} & \textbf{0.036} & \textbf{0.019} & \textbf{0.013} & \textbf{0.208} & \textbf{0.047} & \textbf{0.163} \\
\midrule
\multicolumn{9}{l}{\textit{Natural empathetic reference -- MELD}} \\
Audio Flamingo 3 & Baseline & 0.055 & 0.018 & 0.011 & 0.008 & 0.144 & 0.027 & 0.121 \\
Audio Flamingo 3 & ER-EDF & \textbf{0.058} & \textbf{0.019} & \textbf{0.012} & \textbf{0.009} & \textbf{0.158} & \textbf{0.029} & \textbf{0.134} \\
Gemma-4-E4B-IT & Baseline & 0.054 & 0.021 & 0.013 & 0.010 & 0.150 & 0.030 & 0.127 \\
Gemma-4-E4B-IT & ER-EDF & \textbf{0.056} & \textbf{0.022} & \textbf{0.014} & 0.010 & \textbf{0.156} & \textbf{0.032} & \textbf{0.131} \\
Megrez-3B-Omni & Baseline & 0.067 & 0.022 & 0.013 & 0.010 & 0.156 & 0.028 & 0.132 \\
Megrez-3B-Omni & ER-EDF & \textbf{0.070} & \textbf{0.023} & \textbf{0.014} & \textbf{0.011} & \textbf{0.165} & \textbf{0.029} & \textbf{0.137} \\
MiDashengLM-7B & Baseline & 0.047 & 0.019 & 0.012 & 0.009 & 0.143 & \textbf{0.029} & 0.127 \\
MiDashengLM-7B & ER-EDF & \textbf{0.049} & 0.019 & 0.012 & \textbf{0.010} & \textbf{0.148} & 0.028 & \textbf{0.128} \\
Step-Audio 2 mini & Baseline & 0.057 & 0.022 & 0.013 & 0.009 & 0.149 & 0.034 & 0.127 \\
Step-Audio 2 mini & ER-EDF & \textbf{0.076} & \textbf{0.028} & \textbf{0.016} & \textbf{0.012} & \textbf{0.171} & \textbf{0.038} & \textbf{0.142} \\
\midrule
\multicolumn{9}{l}{\textit{Empathy-centric reference -- IEMOCAP}} \\
Audio Flamingo 3 & Baseline & 0.031 & 0.008 & 0.004 & 0.003 & 0.130 & 0.021 & 0.101 \\
Audio Flamingo 3 & ER-EDF & \textbf{0.037} & \textbf{0.011} & \textbf{0.005} & \textbf{0.004} & \textbf{0.150} & \textbf{0.028} & \textbf{0.117} \\
Gemma-4-E4B-IT & Baseline & 0.024 & 0.008 & 0.004 & 0.003 & 0.159 & 0.030 & 0.125 \\
Gemma-4-E4B-IT & ER-EDF & \textbf{0.030} & \textbf{0.011} & \textbf{0.006} & \textbf{0.004} & \textbf{0.189} & \textbf{0.043} & \textbf{0.147} \\
Megrez-3B-Omni & Baseline & 0.056 & 0.014 & 0.007 & 0.005 & 0.186 & 0.035 & 0.139 \\
Megrez-3B-Omni & ER-EDF & \textbf{0.087} & \textbf{0.027} & \textbf{0.014} & \textbf{0.008} & \textbf{0.242} & \textbf{0.056} & \textbf{0.172} \\
MiDashengLM-7B & Baseline & 0.016 & 0.005 & 0.002 & 0.002 & 0.126 & 0.020 & 0.104 \\
MiDashengLM-7B & ER-EDF & \textbf{0.022} & \textbf{0.006} & \textbf{0.003} & 0.002 & \textbf{0.149} & \textbf{0.024} & \textbf{0.120} \\
Step-Audio 2 mini & Baseline & 0.035 & 0.010 & 0.005 & 0.003 & 0.142 & 0.025 & 0.113 \\
Step-Audio 2 mini & ER-EDF & \textbf{0.073} & \textbf{0.024} & \textbf{0.012} & \textbf{0.008} & \textbf{0.204} & \textbf{0.043} & \textbf{0.152} \\
\midrule
\multicolumn{9}{l}{\textit{Empathy-centric reference -- MELD}} \\
Audio Flamingo 3 & Baseline & \textbf{0.021} & 0.006 & 0.003 & 0.002 & 0.136 & 0.027 & 0.106 \\
Audio Flamingo 3 & ER-EDF & 0.020 & \textbf{0.007} & \textbf{0.004} & 0.002 & \textbf{0.149} & \textbf{0.031} & \textbf{0.118} \\
Gemma-4-E4B-IT & Baseline & 0.011 & 0.004 & 0.002 & 0.002 & 0.135 & 0.026 & 0.107 \\
Gemma-4-E4B-IT & ER-EDF & \textbf{0.012} & 0.004 & \textbf{0.003} & 0.002 & \textbf{0.146} & \textbf{0.031} & \textbf{0.115} \\
Megrez-3B-Omni & Baseline & 0.029 & 0.008 & 0.004 & 0.003 & 0.156 & 0.029 & 0.120 \\
Megrez-3B-Omni & ER-EDF & \textbf{0.036} & \textbf{0.011} & \textbf{0.006} & \textbf{0.004} & \textbf{0.177} & \textbf{0.032} & \textbf{0.134} \\
MiDashengLM-7B & Baseline & 0.008 & 0.003 & 0.001 & 0.001 & 0.109 & 0.017 & 0.091 \\
MiDashengLM-7B & ER-EDF & \textbf{0.009} & 0.003 & \textbf{0.002} & 0.001 & \textbf{0.122} & \textbf{0.019} & \textbf{0.100} \\
Step-Audio 2 mini & Baseline & 0.025 & 0.007 & 0.004 & 0.002 & 0.128 & 0.021 & 0.100 \\
Step-Audio 2 mini & ER-EDF & \textbf{0.038} & \textbf{0.011} & \textbf{0.005} & \textbf{0.004} & \textbf{0.160} & \textbf{0.027} & \textbf{0.120} \\
\bottomrule
\end{tabular}%
}
\caption{Lexical overlap metrics for generated responses evaluated against natural empathetic and empathy-centric references. B-$n$ denotes BLEU-$n$ and R-$n$/R-L denote ROUGE-$n$/ROUGE-L.}
\end{table*}
ER-EDF consistently outperforms the baseline across all five models, both datasets, and both reference types. On IEMOCAP, the improvements are particularly pronounced: for instance, Step-Audio 2 mini achieves a 75\% relative gain in BLEU-1 under the natural reference and a 109\% gain under the empathy-centric reference. The gains are robust across model scales, holding for the smaller Megrez-3B-Omni as well as the larger 7B models. On MELD, ER-EDF maintains a consistent advantage, with only two marginal ties. The stronger improvements under the empathy-centric reference suggest that ER-EDF steers generation toward more empathetically aligned language, rather than merely increasing lexical overlap with generic conversational continuations.

\paragraph{Lexical Diversity and Emotion Adaptivity}
\label{app:mattr-adaptivity}

Table~\ref{tab:appendix_mattr_emotion_adaptivity} reports within-response
lexical richness and emotion adaptivity metrics for baseline and ER-EDF
responses.

\begin{table*}[htbp]
\centering
\scriptsize
\resizebox{\textwidth}{!}{%
\begin{tabular}{llrrrrrrrr}
\toprule
 & & \multicolumn{4}{c}{\textbf{IEMOCAP}} & \multicolumn{4}{c}{\textbf{MELD}} \\
\cmidrule(lr){3-6}\cmidrule(lr){7-10}
\textbf{Model} & \textbf{Setting} & \textbf{MATTR}$\uparrow$ & \textbf{BAcc$_{\mathrm{emb}}$}$\uparrow$ & \textbf{BAcc$_{\mathrm{tfidf}}$}$\uparrow$ & \textbf{$\eta^2{\times}10^2$}$\uparrow$ & \textbf{MATTR}$\uparrow$ & \textbf{BAcc$_{\mathrm{emb}}$}$\uparrow$ & \textbf{BAcc$_{\mathrm{tfidf}}$}$\uparrow$ & \textbf{$\eta^2{\times}10^2$}$\uparrow$ \\
\midrule
Audio Flamingo 3 & Baseline & 0.9393 & \textbf{0.386} & 0.437 & 0.42 & 0.9648 & 0.185 & 0.210 & 0.76 \\
Audio Flamingo 3 & ER-EDF & \textbf{0.9439} & 0.385 & \textbf{0.439} & \textbf{0.53} & \textbf{0.9677} & \textbf{0.195} & \textbf{0.225} & \textbf{0.96} \\
\addlinespace
Gemma-4-E4B-IT & Baseline & 0.9632 & 0.330 & 0.406 & 0.39 & 0.9619 & 0.199 & 0.219 & 0.74 \\
Gemma-4-E4B-IT & ER-EDF & \textbf{0.9767} & \textbf{0.357} & \textbf{0.449} & \textbf{1.20} & \textbf{0.9664} & \textbf{0.205} & \textbf{0.274} & \textbf{1.43} \\
\addlinespace
Megrez-3B-Omni & Baseline & 0.9774 & 0.328 & 0.362 & 0.46 & \textbf{0.9788} & 0.181 & 0.212 & 1.22 \\
Megrez-3B-Omni & ER-EDF & \textbf{0.9785} & \textbf{0.349} & \textbf{0.411} & \textbf{0.91} & 0.9787 & \textbf{0.221} & \textbf{0.282} & \textbf{1.45} \\
\addlinespace
MiDashengLM-7B & Baseline & 0.9602 & 0.344 & 0.409 & 0.76 & 0.9647 & 0.216 & 0.234 & 0.85 \\
MiDashengLM-7B & ER-EDF & \textbf{0.9647} & \textbf{0.372} & \textbf{0.429} & \textbf{1.03} & \textbf{0.9760} & \textbf{0.234} & \textbf{0.264} & \textbf{1.33} \\
\addlinespace
Step-Audio 2 mini & Baseline & 0.8481 & 0.369 & 0.412 & 0.41 & 0.8830 & 0.207 & 0.254 & 0.99 \\
Step-Audio 2 mini & ER-EDF & \textbf{0.8939} & \textbf{0.374} & \textbf{0.461} & \textbf{0.95} & \textbf{0.9324} & \textbf{0.240} & \textbf{0.282} & \textbf{1.38} \\
\bottomrule
\end{tabular}%
}
\caption{Within-response lexical richness and emotion adaptivity
(IEMOCAP $n{=}4{,}508$, MELD $n{=}2{,}610$). MATTR: moving-average
type--token ratio (window 10 words; items where both responses
$\geq$10 words). BAcc: balanced accuracy of recovering the speaker's
gold emotion from the response via logistic regression on sentence
embeddings (all-mpnet-base-v2) or TF-IDF uni/bigrams, dialogue-grouped
5-fold CV (chance: IEMOCAP 0.250, MELD 0.143).
$\eta^2$: sentence-embedding variance explained by gold emotion.
Bold marks the higher value per model--dataset--metric pair.}
\label{tab:appendix_mattr_emotion_adaptivity}
\end{table*}

%% file: latex/appendix/human.tex
\subsection{Human Survey}
\label{app:human}
\paragraph{Human evaluation ethics and recruitment.}
Human evaluation was conducted through an online annotation interface. Participants were randomly recruited from a university participant pool and were compensated for their time according to institutional guidelines. Before participating, annotators were informed about the purpose of the study, the annotation procedure, the type of dialogue data they would evaluate, and how their responses would be used for research. Informed consent was obtained before the annotation task began. All collected judgments were anonymized and analyzed only in aggregate form. The study protocol was reviewed and approved by the relevant institutional ethics review process. To preserve author anonymity during review, identifying details of the institution and approval record are omitted from the anonymous submission.

\paragraph{Dataset validation survey.}

For dataset validation, we sampled coherent multi-turn spoken dialogue segments from both IEMOCAP and MELD. 
We selected 11 contiguous speech-dialogue instances from each of the five IEMOCAP folds, and additionally sampled 11 instances from the MELD validation split and 11 instances from the MELD test split. 
In total, this yielded 77 unique evaluation instances. 
Nine annotators participated in the study, with seven completing the full set of instances. 
Including valid partial annotations from the remaining annotators, we collected 549 human reviews in total. 
Each speech-dialogue instance received approximately 7.1 independent human judgments on average. 
For each instance, annotators first listened to the spoken dialogue context and then ranked three candidate responses from most empathetic to least empathetic: the original dataset response, the natural empathetic response, and the empathy-focused response. 
We report human evaluation results by aggregating all valid judgments across annotators. 
We use \emph{Win Rate} to measure the proportion of cases in which a generated response is ranked above the original dataset response, and \emph{Rank-1 Rate} to measure the proportion of cases in which a generated response is selected as the best among the three candidates.

\begin{table*}[t]
\centering
\small
\begin{tabularx}{\textwidth}{p{0.13\textwidth}X}
\toprule
\textbf{Field} & \textbf{Content} \\
\midrule
Scene & Scene 1 \\

Instruction & Please listen to the context audio, then rank the responses from most empathetic to least empathetic. \\

\midrule
\multicolumn{2}{l}{\textbf{Spoken dialogue context}} \\
\midrule
Turn 1 & Audio segment, 3 seconds. \\
Turn 2 & Audio segment, 1 second. \\
Turn 3 & Audio segment, 3 seconds. \\
Turn 4 & Audio segment, 4 seconds. \\
Turn 5 & Audio segment, 4 seconds. \\

\midrule
\multicolumn{2}{l}{\textbf{Candidate responses}} \\
\midrule
Response \#1 &
Oh, that's incredibly frustrating. It sounds like you're in a really tough spot right now, and I can see why you'd feel stuck. Let's see if we can figure out what your options are. \\

Response \#2 &
I understand this is really frustrating, especially after having your wallet stolen. Let me just check the system here for a moment and see if there are any alternative options for you. \\

Response \#3 &
How am I supposed to get an ID without an ID? How does a person get an ID in the first place? \\

\midrule
Annotation output & Ranking of the three candidate responses from most empathetic to least empathetic. \\

\bottomrule
\end{tabularx}
\caption{Example human evaluation instance for dataset validation. Annotators first listened to the multi-turn spoken dialogue context and then ranked three candidate responses according to empathetic quality.}
\label{tab:human-eval-dataset-example}
\end{table*}

\paragraph{Main-result pairwise survey.}
For the main result evaluation, we conduct a pairwise human preference study between the baseline response and the ER-EDF response. 
For each evaluated model and dialogue instance, annotators first listened to the same multi-turn spoken dialogue context used for generation. 
They were then shown two candidate responses produced by the same LALM backbone under different prompting settings: the baseline prompt and the ER-EDF prompt. 
Annotators were asked to choose which response was more empathetic, considering both contextual relevance and emotional appropriateness. 
Unlike the dataset validation survey, this evaluation uses a pairwise selection format rather than a three-way ranking format. 
We report the human preference rate of ER-EDF against the baseline, and additionally use human agreement to measure how consistently annotators prefer one response over the other.

\begin{table*}[ht]
\centering
\small
\begin{tabularx}{\textwidth}{p{0.13\textwidth}X}
\toprule
\textbf{Field} & \textbf{Content} \\
\midrule
Model & Audio Flamingo 3 \\

Instruction & Please listen to the context audio, then choose which response is more empathetic. \\

\midrule
\multicolumn{2}{l}{\textbf{Spoken dialogue context}} \\
\midrule
Turn 1 & Audio segment, 3 seconds. \\
Turn 2 & Audio segment, 2 seconds. \\
Turn 3 & Audio segment, 2 seconds. \\
Turn 4 & Audio segment, 4 seconds. \\
Turn 5 & Audio segment, 2 seconds. \\

\midrule
\multicolumn{2}{l}{\textbf{Candidate responses}} \\
\midrule
Response \#1 &
I'm not sure, I wasn't there. What did you think of it? \\

Response \#2 &
I'm so sorry to hear that, can you tell me more about what happened? \\

\midrule
Annotation output & Selection of the more empathetic response between the two candidates. \\

\bottomrule
\end{tabularx}
\caption{Example human evaluation instance for the main result comparison. Annotators listened to the multi-turn spoken dialogue context and selected the more empathetic response from a pair of baseline and ER-EDF outputs.}
\label{tab:human-eval-main-example}
\end{table*}

\subsection{Computing Resources}

All experiments were conducted on NVIDIA A100 80GB GPUs. All model inference and evaluation procedures were implemented with the \texttt{ms-swift} framework.

%% file: emnlp2026_draft.bbl
\begin{thebibliography}{41}
\providecommand{\natexlab}[1]{#1}

\bibitem[{Baig and Shete(2025)}]{empathyllama}
Samrin Baig and Prasanna Shete. 2025.
\newblock Lumina: A fine-tuned llama 3.1 model for empathetic psychological support.
\newblock In \emph{2025 12th International Conference on Future Internet of Things and Cloud (FiCloud)}, pages 431--437. IEEE.

\bibitem[{Batson(1991)}]{batson1991altruism}
C.~Daniel Batson. 1991.
\newblock \href {https://doi.org/10.4324/9781315808048} {\emph{The Altruism Question: Toward a Social-Psychological Answer}}.
\newblock Lawrence Erlbaum Associates, Hillsdale, NJ.

\bibitem[{Busso et~al.(2008)Busso, Bulut, Lee, Kazemzadeh, Mower, Kim, Chang, Lee, and Narayanan}]{busso2008iemocap}
Carlos Busso, Murtaza Bulut, Chi-Chun Lee, Abe Kazemzadeh, Emily Mower, Sungbok Kim, Jeannette~N. Chang, Sungbok Lee, and Shrikanth~S. Narayanan. 2008.
\newblock Iemocap: Interactive emotional dyadic motion capture database.
\newblock In \emph{Language Resources and Evaluation}.

\bibitem[{Chamishka et~al.(2022)Chamishka, Madhavi, Nawaratne, Alahakoon, De~Silva, Chilamkurti, and Nanayakkara}]{MELDisshorter}
Sadil Chamishka, Ishara Madhavi, Rashmika Nawaratne, Damminda Alahakoon, Daswin De~Silva, Naveen Chilamkurti, and Vishaka Nanayakkara. 2022.
\newblock A voice-based real-time emotion detection technique using recurrent neural network empowered feature modelling.
\newblock \emph{Multimedia Tools and Applications}.

\bibitem[{Covington and McFall(2010)}]{covington2010mattr}
Michael~A. Covington and Joe~D. McFall. 2010.
\newblock \href {https://doi.org/10.1080/09296171003643098} {Cutting the {G}ordian knot: The moving-average type--token ratio ({MATTR})}.
\newblock \emph{Journal of Quantitative Linguistics}, 17(2):94--100.

\bibitem[{Dinkel et~al.(2025)Dinkel, Li, Liu, Luan, Niu, Sun, Wang, Xiao, Zhang, and Zhou}]{dinkel2025midashenglm}
Heinrich Dinkel, Gang Li, Jizhong Liu, Jian Luan, Yadong Niu, Xingwei Sun, Tianzi Wang, Qiyang Xiao, Junbo Zhang, and Jiahao Zhou. 2025.
\newblock \href {https://arxiv.org/abs/2508.03983} {{MiDashengLM}: Efficient audio understanding with general audio captions}.
\newblock \emph{Preprint}, arXiv:2508.03983.

\bibitem[{Drapeau et~al.(2017)Drapeau, Blake, Dobson, and K{\"o}rner}]{regsad2}
Martin Drapeau, Emily Blake, Keith~S Dobson, and Annett K{\"o}rner. 2017.
\newblock Coping strategies in major depression and over the course of cognitive therapy for depression.
\newblock \emph{Canadian journal of counselling and psychotherapy}, 51(1).

\bibitem[{Fredrickson(2004)}]{regpos}
Barbara~L Fredrickson. 2004.
\newblock The broaden--and--build theory of positive emotions.
\newblock \emph{Philosophical transactions of the royal society of London. Series B: Biological Sciences}, 359(1449):1367--1377.

\bibitem[{{Gemma Team}(2026)}]{gemmateam2026gemma4}
{Gemma Team}. 2026.
\newblock \href {https://arxiv.org/abs/2607.02770} {{Gemma 4} technical report}.
\newblock \emph{Preprint}, arXiv:2607.02770.

\bibitem[{Geng et~al.(2025)Geng, Shao, Xue, Wang, Xie, Guo, Zhao, Li, Tian, Wang et~al.}]{OSUM-Echat}
Xuelong Geng, Qijie Shao, Hongfei Xue, Shuiyuan Wang, Hanke Xie, Zhao Guo, Yi~Zhao, Guojian Li, Wenjie Tian, Chengyou Wang, and 1 others. 2025.
\newblock Osum-echat: Enhancing end-to-end empathetic spoken chatbot via understanding-driven spoken dialogue.
\newblock \emph{arXiv preprint arXiv:2508.09600}.

\bibitem[{Goel et~al.(2025)Goel, Ghosh, Kim, Kumar, Kong, gil Lee, Yang, Duraiswami, Manocha, Valle, and Catanzaro}]{goel2025audioflamingo3}
Arushi Goel, Sreyan Ghosh, Jaehyeon Kim, Sonal Kumar, Zhifeng Kong, Sang gil Lee, Chao-Han~Huck Yang, Ramani Duraiswami, Dinesh Manocha, Rafael Valle, and Bryan Catanzaro. 2025.
\newblock \href {https://arxiv.org/abs/2507.08128} {{Audio Flamingo 3}: Advancing audio intelligence with fully open large audio language models}.
\newblock \emph{Preprint}, arXiv:2507.08128.

\bibitem[{{Google DeepMind}(2025)}]{google2024gemini}
{Google DeepMind}. 2025.
\newblock \href {https://arxiv.org/abs/2507.06261} {Gemini 2.5: Pushing the frontier with advanced reasoning, multimodality, long context, and next generation agentic capabilities}.
\newblock \emph{Preprint}, arXiv:2507.06261.

\bibitem[{Gross(1998)}]{gross1998emerging}
James~J. Gross. 1998.
\newblock The emerging field of emotion regulation: An integrative review.
\newblock \emph{Review of General Psychology}, 2(3):271--299.

\bibitem[{Gross(2015)}]{Gross2015-bk}
James~J Gross. 2015.
\newblock The extended process model of emotion regulation: Elaborations, applications, and future directions.
\newblock \emph{Psychol. Inq.}, 26(1):130--137.

\bibitem[{Kim and Vossen(2021)}]{kim2021emoberta}
Taewoon Kim and Piek Vossen. 2021.
\newblock Emoberta: Speaker-aware emotion recognition in conversation with roberta.
\newblock \emph{arXiv preprint arXiv:2108.12009}.

\bibitem[{Laranjo et~al.(2018)Laranjo, Dunn, Tong, Kocaballi, Chen, Bashir, Surian, Gallego, Magrabi, Lau, and Coiera}]{laranjo2018conversational}
Liliana Laranjo, Adam~G. Dunn, Huong~Ly Tong, Ahmet~Baki Kocaballi, Jessica Chen, Rabia Bashir, Didi Surian, Blanca Gallego, Farah Magrabi, Annie Y.~S. Lau, and Enrico Coiera. 2018.
\newblock \href {https://doi.org/10.1093/jamia/ocy072} {Conversational agents in healthcare: a systematic review}.
\newblock \emph{Journal of the American Medical Informatics Association}, 25(9):1248--1258.

\bibitem[{Li et~al.(2025)Li, Li, Li, Liu, Liu, Niu, Tan, Xu, Yao, Yuan, Zhou, Zhuang, Yan, Dai, and Wang}]{li2025megrez}
Boxun Li, Yadong Li, Zhiyuan Li, Congyi Liu, Weilin Liu, Guowei Niu, Zheyue Tan, Haiyang Xu, Zhuyu Yao, Tao Yuan, Dong Zhou, Yueqing Zhuang, Shengen Yan, Guohao Dai, and Yu~Wang. 2025.
\newblock \href {https://arxiv.org/abs/2502.15803} {Megrez-omni technical report}.
\newblock \emph{Preprint}, arXiv:2502.15803.

\bibitem[{Li et~al.(2022)Li, Li, Ren, Ren, and Chen}]{li2022kemp}
Qintong Li, Piji Li, Zhaochun Ren, Pengjie Ren, and Zhumin Chen. 2022.
\newblock \href {https://doi.org/10.1609/aaai.v36i10.21347} {Knowledge bridging for empathetic dialogue generation}.
\newblock In \emph{Proceedings of the AAAI Conference on Artificial Intelligence}, volume~36, pages 10993--11001.

\bibitem[{Lin et~al.(2024{\natexlab{a}})Lin, Chiang, and Lee}]{lin2024advancing}
Guan-Ting Lin, Cheng-Han Chiang, and Hung-yi Lee. 2024{\natexlab{a}}.
\newblock \href {https://doi.org/10.18653/v1/2024.acl-long.358} {Advancing large language models to capture varied speaking styles and respond properly in spoken conversations}.
\newblock In \emph{Proceedings of the 62nd Annual Meeting of the Association for Computational Linguistics (Volume 1: Long Papers)}, pages 6626--6642. Association for Computational Linguistics.

\bibitem[{Lin et~al.(2024{\natexlab{b}})Lin, Shivakumar, Gandhe, Yang, Gu, Ghosh, Stolcke, Lee, and Bulyko}]{lin2024paralingpt}
Guan-Ting Lin, Prashanth~Gurunath Shivakumar, Ankur Gandhe, Chao-Han~Huck Yang, Yile Gu, Shalini Ghosh, Andreas Stolcke, Hung-yi Lee, and Ivan Bulyko. 2024{\natexlab{b}}.
\newblock \href {https://doi.org/10.1109/ICASSP48485.2024.10446933} {Paralinguistics-enhanced large language modeling of spoken dialogue}.
\newblock In \emph{ICASSP 2024 - 2024 IEEE International Conference on Acoustics, Speech and Signal Processing (ICASSP)}, pages 10316--10320. IEEE.

\bibitem[{Liu et~al.(2021)Liu, Zheng, Demasi, Sabour, Li, Yu, Jiang, and Huang}]{liu-etal-2021-towards}
Siyang Liu, Chujie Zheng, Orianna Demasi, Sahand Sabour, Yu~Li, Zhou Yu, Yong Jiang, and Minlie Huang. 2021.
\newblock \href {https://doi.org/10.18653/v1/2021.acl-long.269} {Towards emotional support dialog systems}.
\newblock In \emph{Proceedings of the 59th Annual Meeting of the Association for Computational Linguistics and the 11th International Joint Conference on Natural Language Processing (Volume 1: Long Papers)}, pages 3469--3483, Online. Association for Computational Linguistics.

\bibitem[{Majumder et~al.(2020)Majumder, Hong, Peng, Lu, Ghosal, Gelbukh, Mihalcea, and Poria}]{majumder2020mime}
Navonil Majumder, Pengfei Hong, Shanshan Peng, Jiankun Lu, Deepanway Ghosal, Alexander Gelbukh, Rada Mihalcea, and Soujanya Poria. 2020.
\newblock \href {https://doi.org/10.18653/v1/2020.emnlp-main.721} {{MIME}: {MIM}icking emotions for empathetic response generation}.
\newblock In \emph{Proceedings of the 2020 Conference on Empirical Methods in Natural Language Processing (EMNLP)}, pages 8968--8979. Association for Computational Linguistics.

\bibitem[{Niven et~al.(2009)Niven, Totterdell, and Holman}]{niven2009classification}
Karen Niven, Peter Totterdell, and David Holman. 2009.
\newblock \href {https://doi.org/10.1037/a0015962} {A classification of controlled interpersonal affect regulation strategies}.
\newblock \emph{Emotion}, 9(4):498--509.

\bibitem[{OpenAI(2024)}]{openai2024gpt4o}
OpenAI. 2024.
\newblock \href {https://arxiv.org/abs/2410.21276} {Gpt-4o system card}.
\newblock \emph{Preprint}, arXiv:2410.21276.

\bibitem[{Poria et~al.(2019)Poria, Hazarika, Majumder, Naik, Cambria, and Mihalcea}]{li2018meld}
Soujanya Poria, Devamanyu Hazarika, Navonil Majumder, Gautam Naik, Erik Cambria, and Rada Mihalcea. 2019.
\newblock Meld: A multimodal multi-party dataset for emotion recognition in conversations.
\newblock In \emph{Proceedings of ACL}.

\bibitem[{Posner et~al.(2005)Posner, Russell, and Peterson}]{reganger}
Jonathan Posner, James~A Russell, and Bradley~S Peterson. 2005.
\newblock The circumplex model of affect: An integrative approach to affective neuroscience, cognitive development, and psychopathology.
\newblock \emph{Development and psychopathology}, 17(3):715--734.

\bibitem[{Preston and de~Waal(2002)}]{preston2002empathy}
Stephanie~D. Preston and Frans B.~M. de~Waal. 2002.
\newblock \href {https://doi.org/10.1017/S0140525X02000018} {Empathy: Its ultimate and proximate bases}.
\newblock \emph{Behavioral and Brain Sciences}, 25(1):1--20.

\bibitem[{Rashkin et~al.(2019)Rashkin, Smith, Li, and Boureau}]{rashkin2019empathetic}
Hannah Rashkin, Eric~Michael Smith, Margaret Li, and Y-Lan Boureau. 2019.
\newblock \href {https://doi.org/10.18653/v1/P19-1534} {Towards empathetic open-domain conversation models: A new benchmark and dataset}.
\newblock In \emph{Proceedings of the 57th Annual Meeting of the Association for Computational Linguistics (ACL)}, pages 5370--5381. Association for Computational Linguistics.

\bibitem[{Reimers and Gurevych(2019)}]{reimers2019sentencebert}
Nils Reimers and Iryna Gurevych. 2019.
\newblock \href {https://doi.org/10.18653/v1/D19-1410} {Sentence-{BERT}: Sentence embeddings using {S}iamese {BERT}-networks}.
\newblock In \emph{Proceedings of the 2019 Conference on Empirical Methods in Natural Language Processing and the 9th International Joint Conference on Natural Language Processing (EMNLP-IJCNLP)}, pages 3982--3992, Hong Kong, China. Association for Computational Linguistics.

\bibitem[{Richardson(2011)}]{richardson2011eta}
John T.~E. Richardson. 2011.
\newblock \href {https://doi.org/10.1016/j.edurev.2010.12.001} {Eta squared and partial eta squared as measures of effect size in educational research}.
\newblock \emph{Educational Research Review}, 6(2):135--147.

\bibitem[{Russell(1980)}]{russell1980circumplex}
James~A. Russell. 1980.
\newblock A circumplex model of affect.
\newblock \emph{Journal of Personality and Social Psychology}, 39(6):1161--1178.

\bibitem[{Sedoc et~al.(2020)Sedoc, Buechel, Nachmany, Buffone, and Ungar}]{WWBP}
Jo{\~a}o Sedoc, Sven Buechel, Yehonathan Nachmany, Anneke Buffone, and Lyle Ungar. 2020.
\newblock \href {https://aclanthology.org/2020.lrec-1.206/} {Learning word ratings for empathy and distress from document-level user responses}.
\newblock In \emph{Proceedings of the Twelfth Language Resources and Evaluation Conference}, pages 1664--1673, Marseille, France. European Language Resources Association.

\bibitem[{Sharma et~al.(2020)Sharma, Miner, Atkins, and Althoff}]{sharma2020computational}
Ashish Sharma, Adam~S. Miner, David~C. Atkins, and Tim Althoff. 2020.
\newblock \href {https://doi.org/10.18653/v1/2020.emnlp-main.425} {A computational approach to understanding empathy expressed in text-based mental health support}.
\newblock In \emph{Proceedings of the 2020 Conference on Empirical Methods in Natural Language Processing (EMNLP)}, pages 5263--5276. Association for Computational Linguistics.

\bibitem[{Timonen and Liukkonen(2008)}]{regsad1}
Markku Timonen and Timo Liukkonen. 2008.
\newblock Management of depression in adults.
\newblock \emph{Bmj}, 336(7641):435--439.

\bibitem[{Wang et~al.(2024{\natexlab{a}})Wang, Liao, Huang, Wu, Zong, and Zhang}]{wang2024blspemo}
Chen Wang, Minpeng Liao, Zhongqiang Huang, Junhong Wu, Chengqing Zong, and Jiajun Zhang. 2024{\natexlab{a}}.
\newblock {BLSP-Emo}: Towards empathetic large speech-language models.
\newblock In \emph{Proceedings of the 2024 Conference on Empirical Methods in Natural Language Processing (EMNLP)}. Association for Computational Linguistics.

\bibitem[{Wang et~al.(2024{\natexlab{b}})Wang, Liao, Huang, Wu, Zong, and Zhang}]{BLSP-Emo}
Chen Wang, Minpeng Liao, Zhongqiang Huang, Junhong Wu, Chengqing Zong, and Jiajun Zhang. 2024{\natexlab{b}}.
\newblock Blsp-emo: Towards empathetic large speech-language models.
\newblock In \emph{Proceedings of the 2024 Conference on Empirical Methods in Natural Language Processing}, pages 19186--19199.

\bibitem[{Wu et~al.(2025)Wu, Yan, Hu, Yi, Feng, Tian, Shen, Yu et~al.}]{wu2025stepaudio2}
Boyong Wu, Chao Yan, Chen Hu, Cheng Yi, Chengli Feng, Fei Tian, Feiyu Shen, Gang Yu, and 1 others. 2025.
\newblock \href {https://arxiv.org/abs/2507.16632} {{Step-Audio 2} technical report}.
\newblock \emph{Preprint}, arXiv:2507.16632.

\bibitem[{Xue et~al.(2024)Xue, Liang, Mu, Zhang, Chen, Chen, and Xie}]{xue2024echat}
Hongfei Xue, Yuhao Liang, Bingshen Mu, Shiliang Zhang, Mengzhe Chen, Qian Chen, and Lei Xie. 2024.
\newblock \href {https://doi.org/10.1109/ISCSLP63861.2024.10800447} {{E-Chat}: Emotion-sensitive spoken dialogue system with large language models}.
\newblock In \emph{Proceedings of the 14th International Symposium on Chinese Spoken Language Processing (ISCSLP)}, pages 586--590. IEEE.

\bibitem[{Zaki and Williams(2013)}]{zaki2013interpersonal}
Jamil Zaki and W~Craig Williams. 2013.
\newblock Interpersonal emotion regulation.
\newblock \emph{Emotion}, 13(5):803.

\bibitem[{Zhang et~al.(2020)Zhang, Kishore, Wu, Weinberger, and Artzi}]{zhang2020bertscore}
Tianyi Zhang, Varsha Kishore, Felix Wu, Kilian~Q. Weinberger, and Yoav Artzi. 2020.
\newblock {BERTScore}: Evaluating text generation with {BERT}.
\newblock \emph{International Conference on Learning Representations}.

\bibitem[{Zheng et~al.(2023)Zheng, Chiang, Sheng, Zhuang, Wu, Zhuang, Lin, Li, Li, Xing, Gonzalez, Stoica, and Zhang}]{zheng2023judging}
Lianmin Zheng, Wei-Lin Chiang, Ying Sheng, Siyuan Zhuang, Zhanghao Wu, Yonghao Zhuang, Zi~Lin, Zhuohan Li, Dacheng Li, Eric~P. Xing, Joseph~E. Gonzalez, Ion Stoica, and Hao Zhang. 2023.
\newblock Judging {LLM}-as-a-judge with {MT}-bench and chatbot arena.
\newblock \emph{arXiv preprint arXiv:2306.05685}.

\end{thebibliography}
